\RequirePackage{fix-cm}

\documentclass[smallcondensed]{svjour3}     

\AtBeginDocument{%
	\paperwidth=\dimexpr
	1in + \oddsidemargin
	+ \textwidth
	+ 1in + \oddsidemargin
	\relax
	\paperheight=\dimexpr
	1in + \topmargin
	+ \headheight + \headsep
	+ \textheight
	+ 1in + \topmargin
	\relax
	\usepackage[pass]{geometry}\relax
}

\smartqed  
\usepackage{subfig}
\usepackage{graphicx}
\usepackage{amsmath}
\usepackage{url}
\usepackage[linesnumbered,ruled]{algorithm2e}
\usepackage[sort&compress,numbers]{natbib}
\usepackage[colorlinks,allcolors=blue]{hyperref}

\usepackage{pgfplots}
\usepackage{multirow}

\newcommand{\bigO}{\mathcal{O}}

\begin{document}
\def\makeheadbox{\relax}
\title{Text Mining using Nonnegative Matrix Factorization and Latent Semantic Analysis}

\author{Ali Hassani \and Amir Iranmanesh \and Najme Mansouri}


\institute{
			A. Hassani \at
			Department of Computer Science, Shahid Bahonar University of Kerman, 76169-14111, Pajoohesh Square, Kerman, I.R. Iran. \\
			\email{alihassanijr1998@gmail.com}
			\and
			A. Iranmanesh \at
			Department of Computer Science, Shahid Bahonar University of Kerman, 76169-14111, Pajoohesh Square, Kerman, I.R. Iran. \\
			\email{amir.ir1374@gmail.com}
			\and
			N. Mansouri (Corresponding author) \at
			Department of Computer Science, Shahid Bahonar University of Kerman, 76169-14111, Pajoohesh Square, Kerman, I.R. Iran. \\
			\email{najme.mansouri@gmail.com}
			\\
}

\date{February 2020}

\maketitle{\relax}

\begin{abstract}
Text clustering is arguably one of the most important topics in modern data mining. Nevertheless, text data require tokenization which usually yields a very large and highly sparse term-document matrix, which is usually difficult to process using conventional machine learning algorithms. Methods such as Latent Semantic Analysis have helped mitigate this issue, but are nevertheless not completely stable in practice. As a result, we propose a new feature agglomeration method based on Nonnegative Matrix Factorization, which is employed to separate the terms into groups, and then each group's term vectors are agglomerated into a new feature vector. Together, these feature vectors create a new feature space much more suitable for clustering. In addition, we propose a new deterministic initialization for spherical K-Means, which proves very useful for this specific type of data. In order to evaluate the proposed method, we compare it to some of the latest research done in this field, as well as some of the most practiced methods. In our experiments, we conclude that the proposed method either significantly improves clustering performance, or maintains the performance of other methods, while improving stability in results.
	\keywords{Nonnegative Matrix Factorization \and Text Clustering \and Latent Semantic Analysis \and Dimensionality reduction}
\end{abstract}

\section{Introduction}
\label{sec:intro}
Due to the technical advances in computer science, text mining is a widely studied area with practically many applications. Text mining can be best described as the process of extracting information from a pool of documents. Nowadays, with the ever-growing online data generation by IoT (Internet of Things), the need for suitable processing is also growing. Gartner has estimated around 4.9 billion online devices around the world, with a projection of an increase up to 25 billion by 2020 \cite{xie2019novel}.
A considerable percentage of the data available online consists of websites, blogs, journals, social networks and the like, all of which include a great amount of text. This massive amount of data cannot in its current form be processed by human beings or conventional data processing, leading the world towards improvement and research in data science. Therefore, just like any other form of data, text data also require preprocessing in order to become truly visible to their demographics. This has been one of the main reasons that search engines have become a necessity. Text mining has been widely used in many different areas such as biomedicine \cite{krallinger2005text,zhu2013biomedical}, recommendation systems \cite{davoodi2013semantic}, and intrusion detection in web applications \cite{adeva2007intrusion}. As a form of data mining, this process requires preprocessing, model learning, and evaluation. The information that is extracted depends on the type of preprocessing and model learning used.
Text mining's most known challenges are related to algorithms or languages. The former includes processing and computational challenges such as problems with large-scale and noisy data. It is worth noting that an increase in the number of documents can lead to a much more significant increase in the number of features and therefore call upon “the curse of dimensionality”. The latter however depends on the method, which transforms the texts into a vectorized version, in other words, deciding to select which words or topics to represent in the matrix.
Many different learning methods have been employed in text mining that help to extract useful information. One of the most frequently used methods is text clustering, which separates different texts into a number of groups, named clusters. Text clustering has been applied to SMS topic detection \cite{lin2016topic}, scientific text grouping using citation contexts \cite{aljaber2010document} and web search engines \cite{modha2004clustering}. K-Means, as one of the most frequently applied clustering methods has also been used for text clustering. Nevertheless, dependence on suitable initialization and being limited to partitions have been its greatest weaknesses. Therefore, other clustering methods such as density-based and hierarchical clustering methods have been employed. Nonetheless, Steinbach et al. \cite{steinbach2000comparison} compared the results of hierarchical clustering measures and bisecting K-Means based measures in terms of F-Measure and entropy, and concluded that bisecting K-Means performs better. Text clustering has also faced many new challenges over the years, as new methods for effective text clustering continue to emerge \cite{forsati2013efficient,janani2019text,zhang2010text,lee2003multilingual,qiang2018short}. Janani et al. \cite{janani2019text} proposed a spectral clustering method which relies on Particle Swarm Optimization instead of the regular K-Means clustering. Forsati et al. \cite{forsati2013efficient} on the other hand proposed stochastic algorithms for document clustering.
However, many of the previously proposed methods do not take the dimensionality of the vectorized text data into account, and many others use evolutionary algorithms, which can lead to instabilities as they are stochastic in nature. In order to address these issues, we propose a new method which results in dimensionality reduction by agglomerating of the term vectors from the term-document matrix in order create a new feature space. This can further increase learning performance.
The primary contributions of this approach are as follows:
\begin{enumerate}
	\item It uses Latent Semantic Analysis of the term-document matrix to initialize a Nonnegative Matrix Factorization of that matrix.
	\item It then employs the resulting Nonnegative Matrix Factorization to partition the terms vectors (feature vectors) into different groups.
	\item Afterwards, it agglomerates each group of term vectors into a new feature vector, changing its overall representation, as opposed to LSA and other SVD-based methods which effectively attempt to keep the original representation by projecting the data into lower dimensions.
	\item The method then creates a nearest-neighbors graph of the new feature space in order to find the initial centroids for seeding K-Means.
	\item Finally, it clusters the documents which exist in the newly created feature space using K-Means.
\end{enumerate}
The remainder of this paper is organized as follows. Section \ref{sec:related} covers related research done in this area. Section \ref{sec:background} provides a detailed background of the necessary concepts. Section \ref{sec:method} describes the proposed feature agglomeration and clustering algorithm. Sections \ref{sec:experiments} and \ref{sec:discussion} present the detailed discussion on results and observations. Section \ref{sec:conclusion} indicates the conclusion of the proposed strategy and future directions.

\section{Related Work}
\label{sec:related}
Research in text clustering has gained considerable attention in the past few years.
Thakran et al. \cite{thakran2014novel} proposed a novel hierarchical agglomerative clustering algorithm, which uses a “Cluster spread” as the linkage metric for agglomeration and clustering threshold. Their method however processes datasets without any special representation or dimensionality reduction, which can be troublesome in text clustering, as the dimensionality of the vectorized documents in text mining is considerably high. Combining the high dimensionality with agglomerative clustering which requires continuous distance computation regardless of the linkage, can be of great computational burden. Karaa et al. \cite{karaa2016medline} also proposed a Genetic Algorithm optimization, which is initialized using an agglomerative clustering tree of the medical MEDLINE dataset. The fitness function of this method is the objective function for clustering. Putting aside the lack of dimensionality reduction, the method requires the computation of the entire agglomeration tree, which can be time-consuming.

Garg et al. \cite{garg2018performance} on the other hand proposed a Genetic-based K-Means centroid initialization, the fitness of which is based on cluster inter- and outer-cluster similarity. Nevertheless, this method is also prone to one of the biggest problems in text mining, which is high dimensionality and sparsity. Lack of feature space reduction or lack of change in representation with center-based clustering can be pointed out as the greatest weaknesses of this method. Janani et al. \cite{janani2019text} on the other hand proposed a Particle Swarm Optimization for a center-based clustering on top of a spectral embedding using nearest-neighbors graph. This approach reduces dimensionality and generates a smaller and better-represented feature space. However, the main weakness of this method is the use of evolutionary optimization, which is stochastic in nature and may lead to instability in results in certain cases. Moreover, the fitness function is set to the objective function, which has a complexity of $\bigO(nK)$ ($K$ being the number of clusters, $n$ being the number of records) and is run per each particle per each iteration, which can be very time-consuming and even impractical in some cases. Gulnashin et al. \cite{gulnashin2019new} proposed an improvement to another novel spherical K-Means initialization, which computes an initial location for spherical K-Means clustering. The improvement includes making the former method less deterministic in order to avoid overlapping centroids and therefore empty clusters. While both perform rapidly in centroid precomputation, their weakness is again the same as all center-based methods. The problem of high dimensionality in text mining can also be pointed out as another weakness of these methods.

Kushwaha et al. \cite{kushwaha2018link} proposed a link-based binary PSO optimization for feature selection on the vectorized texts and from there applied K-Means for text clustering. In order to address the high dimensionality of tokenized text data, the authors of this paper added a feature selection step to this algorithm. However effective, feature selection methods can sometimes be more time consuming than conventional matrix methods, such as Principle Component Analysis (PCA), Spectral Embedding, and Nonnegative Matrix Factorization (NMF). Nevertheless, while center-based clustering may be suitable for big data when considering the lower computational costs, their main weakness is the inability to cluster data with varying densities.

Revanasiddappa et al. \cite{revanasiddappa2018clustering} proposed a kernel possibilistic model of fuzzy C-Means in order to make the original C-Means algorithm less sensitive to noise, and improves the classical possibilistic model by using a kernel distance metric. In this method, the kernel representation can further improve the clustering, along with the possibilistic C-Means.  Nevertheless, like many others, the high dimensionality can be pointed out as this method's most obvious weakness.

Ahmadi et al. \cite{ahmadi2018cluster} used a Sparse Topical Coding-based method, which takes advantage of bag of words models and topic space projections in order to improve text clustering. This projection is essentially a change in the feature space, which can further improve clustering results.

An overview on the related methods is presented in Table \ref{tab:relatedwork}.
\begin{table}
	\centering
	\caption{An overview of previous studies in text clustering}
	\label{tab:relatedwork}
	\begin{tabular}{p{0.025\textwidth}p{0.15\textwidth}p{0.1\textwidth}p{0.11\textwidth}p{0.16\textwidth}p{0.24\textwidth}}
		\hline\noalign{\smallskip}
		Year&Reference&Field&Datasets&Clustering& Dim. Red.\\
		\noalign{\smallskip}\hline\noalign{\smallskip}
		\citeyear{thakran2014novel}&\citet{thakran2014novel}&Medical&Liver Disorder, Heart&Agglomerative&None\\
		
		\citeyear{karaa2016medline}&\citet{karaa2016medline}&Medical&MEDLINE&Agglomerative + GA&None\\
		
		\citeyear{kushwaha2018link}&\citet{kushwaha2018link}&Big data&TDT2, Reuters&Center-based + BPSO&Feature Selection\\
		
		\citeyear{garg2018performance}&\citet{garg2018performance}&General&Classic, 20News&Center-based + GA&None\\
		
		\citeyear{janani2019text}&\citet{janani2019text}&General&Reuters, TDT2&Center-based + PSO&Spectral Embedding\\
		
		\citeyear{ahmadi2018cluster}&\citet{ahmadi2018cluster}&General&20News, WebKB&Center-based&Sparse Topical Coding\\
		
		\citeyear{revanasiddappa2018clustering}&\citet{revanasiddappa2018clustering}&General&20News&Fuzzy Center-based&Kernel representation\\
		
		\citeyear{gulnashin2019new}&\citet{gulnashin2019new}&General&Reuters, 20News&Center-based&None\\
		
		\noalign{\smallskip}\hline
	\end{tabular}
\end{table}
As it can be observed, over half of the methods presented either have no dimensionality reduction or increase the dimensionality of the already high-dimensional text data, which can be troublesome in practical cases, while it may produce slightly better results with better representation. Moreover, half of the methods use evolutionary algorithms, which may lead to instabilities in results, and high computational costs per iteration. Another notable fact is that some have used agglomerative clustering methods, which include a rather considerable computational burden. In the proposed approach, we sought out to decrease dimensionality and change the feature representation at the same time by using Nonnegative Matrix Factorization (NMF), on top of introducing a deterministic K-Means initialization in order to maintain stability.

\section{Background Information}
\label{sec:background}
In this section, we briefly review the basic information about text tokenization, term weighting, eigen-decomposition, singular value decomposition, nonnegative matrix factorization, latent semantic analysis, and nearest-neighbors graph.
\subsection{Text Tokenization}
\begin{figure}
	\centering
	\includegraphics[width=0.9\textwidth]{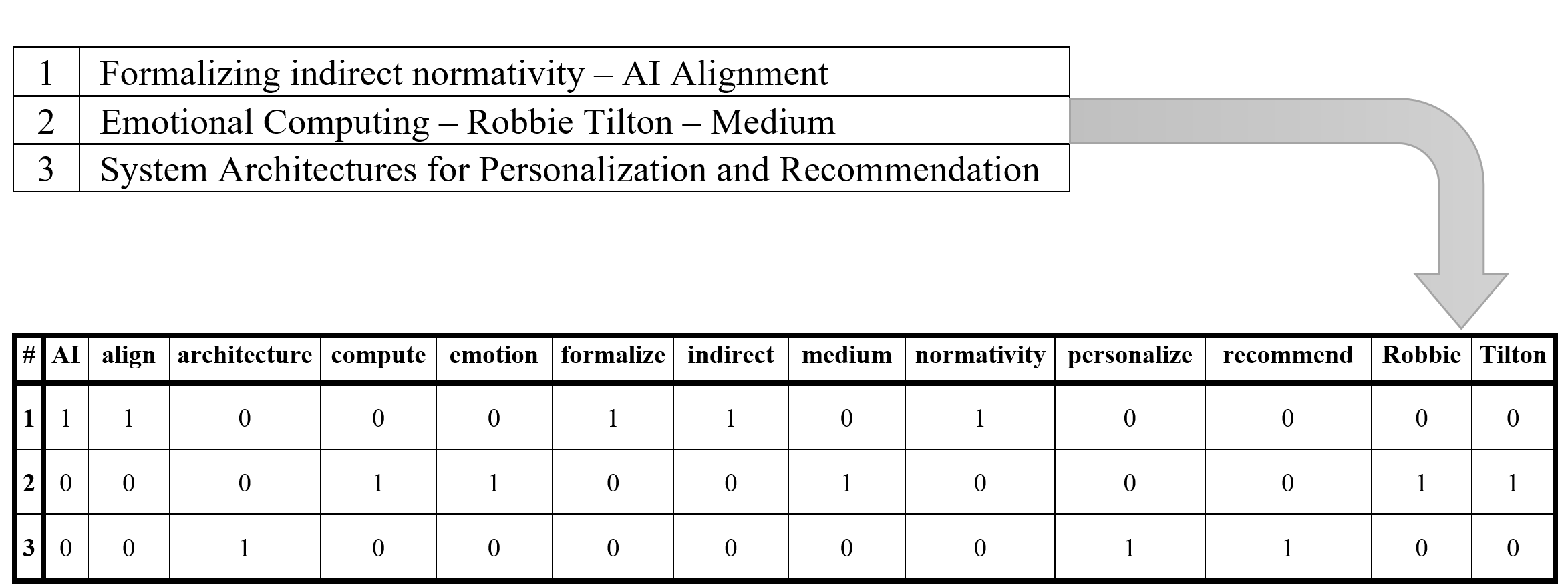}
	\caption{An example of the tokenization of three text strings}
	\label{fig:tokenization}
\end{figure}
The general purpose of text mining is processing a set of texts such as article titles, text messages and the like called documents. In order to process this information, a representation of these texts is required. Tokenization in text mining is the process which creates a vectorized representation for text data. This process segments large texts into sentences and the sentences are then tokenized into words.
Figure \ref{fig:tokenization} depicts a sample of the tokenization of three text strings from the Medium.com text dataset obtained from Kaggle \cite{sankesara2018}. As it can be seen, the word “for” is excluded. In text tokenization such words, called stop words are excluded. Another process that can be included in this section is called stemming, which reduces words to their stems (e.g., Computing to compute).

\subsection{Term weighting}
In this paper, we use one of the most frequently used term weighting methods, TF-IDF \cite{sparck1972statistical} (the product of Term Frequency and Inverse Document Frequency).
Given the matrix $R = [ r_{ij} ] \in Z_{t \times n}$ which represents the count of the occurrences of each term in each document, Term and Inverse Document Frequencies are expressed in Eq. \eqref{eq:tfidf}.
\begin{flalign}
\label{eq:tfidf}
	\begin{aligned}
	&TF({Term}_i,{Document}_j) = \log(1+r_{ij})\\
	&IDF({Term}_i) = \log(\frac{n}{n_{{Term}_i}})\\
	&\text{where } n_{{Term}_i} \text{ is the number of documents containing } {Term}_i\text{.}
	\end{aligned}
\end{flalign}
This method applies weights to the terms based on their frequency in the documents, therefore yielding a more accurate representation of the text documents.
TF-IDF has been widely used in text mining as it is one of the most popular and effective term weighting methods \cite{janani2019text,karaa2016medline}.

\subsection{Eigen-decomposition and Singular Value Decomposition}
Eigen-decomposition is a matrix factorization method in which diagonalizable matrices can be represented as:
\begin{flalign}
\label{eq:eigen}
	\begin{aligned}
	& A = Q D Q^{-1}\\
	&\text{where } D \text{ is diagonal.}
	\end{aligned}
\end{flalign}
$A$ is diagonalizable if it is a square matrix for which there exists a matrix such as $Q$ where $Q^{-1} AQ $ is a diagonal matrix. The columns of the matrix $Q$ in eigen-decomposition represent the eigenvectors of the matrix.
Eigen-decomposition has been widely applied in machine learning such as Spectral Clustering. This clustering approach operates by creating a record-to-record similarity matrix, computing the Laplacian of that matrix and then computing the eigenvectors of the Laplacian. Afterwards, the vectors are sorted in ascending order by their corresponding eigenvalues, and starting from the second vector, the algorithm chooses a specific number of those vectors to create the new feature space. This clustering approach has become very popular since Shi et al. applied it to image segmentation \cite{shi2000normalized}.
Singular Value Decomposition (SVD) on the other hand is a factorization of any complex or real matrix, which is not contingent on the matrix being in square form. It decomposes the matrix into the product of three matrices:
\begin{equation}
\label{eq:singulardecomp}
X_{n \times m} = U_{n \times n} \ \Sigma_{n \times m} \ V_{m \times m}^T 
\end{equation}
In this decomposition, $U$ and $V$ contain the left and right singular vectors of $X$, respectively, and $\Sigma$ is a rectangular diagonal matrix containing the singular values. The left singular vectors and the right singular vectors are orthonormal matrices as well.
SVD is used in many dimension reduction methods such as Latent Semantic Analysis (LSA) \cite{dumais2004latent} which is probably the most widely applied method for text data \cite{wang2011text,wild2007investigating,yu2008latent,yu2009combining}. LSA increases clustering performance with very little computational burden. Moreover, SVD can also be used for Principal Component Analysis (PCA), which is yet another dimension reduction method widely applied across many fields of machine learning, such as clustering \cite{cohen2015dimensionality,ding2004k,korenius2007principal}.

\subsection{Nonnegative Matrix Factorization}
Nonnegative Matrix Factorization (NMF) is a matrix analysis method, which attempts to represent each matrix in the following format:
\begin{equation}
\label{eq:nmf}
X_{n \times m} = W_{n \times k} \  H_{k \times m}  
\end{equation}
This representation requires an optimization, which is aimed at minimizing the following expression:
\begin{equation}
\label{eq:nmfoptimization}
\| X - W H \|_{F}
\end{equation}
The initialization of the two matrices $W$ and $H$ can be done randomly, but may yield different results each time which can be a problem when applying this method to machine learning. Nevertheless, many proposed seeding methods for NMF. Boutsidis et al. \cite{boutsidis2008svd} proposed a method in which SVD can be used as an initializer for NMF. In their method, a singular value decomposition will yield three matrices which can be processed into two nonnegative matrices which are used as the initial values of $W$ and $H$. Casalino et al. \cite{casalino2014subtractive} on the other hand proposed subtractive clustering for NMF initialization.
NMF itself has been widely applied in many fields in machine learning, especially in clustering \cite{pompili2014two,zeng2014image}. Moreover, research into NMF being used in deep learning has also gained interest \cite{flenner2017deep}. NMF has also been previously applied to biomedical document clustering \cite{huang2011enhanced} as well as semi-supervised document clustering \cite{lu2016semi}.

\subsection{Latent Semantic Analysis}
Latent Semantic Analysis (LSA) \cite{dumais2004latent} is basically a process based on singular value decomposition, which has been widely applied to text mining \cite{song2010latent,wang2009text,zheng2013text}. LSA decomposes a tokenized text data matrix, which usually has a great level of sparsity and uses a rank $k$ approximation by selecting $k$ of the left-singular vectors corresponding to the $k$ largest singular values. This method generates a new space which emboldens the significance in difference between documents and therefore will increase learning performance. The output from LSA is also usually normalized, which essentially maps the documents onto the k-dimensional hyper-sphere.
The new space generated by LSA is computed from the product of the matrices $U_k$ and $\Sigma_k$ where:
\begin{flalign}
\label{eq:lsa}
	\begin{aligned}
	&U_k = [ U_{ij} ],\quad i \in \{ 1,2,…,n \}, \quad j \in \{ 1,2,…,k \}\\
	&\Sigma_k = [ \Sigma_{ij} ], \quad i \in \{ 1,2,…,k \}, \quad j \in \{ 1,2,…,k \}\\
	&\text{where } k \text{ is the number of vectors selected.}\\
	&X_{LSA} = U_k \ \Sigma_k  
	\end{aligned}
\end{flalign}
Afterwards, any learning algorithm can be fit on the new projected space, $X_{LSA}$.
\subsection{Nearest-neighbors graph}
\begin{figure}
	\centering
	\includegraphics[width=0.55\textwidth]{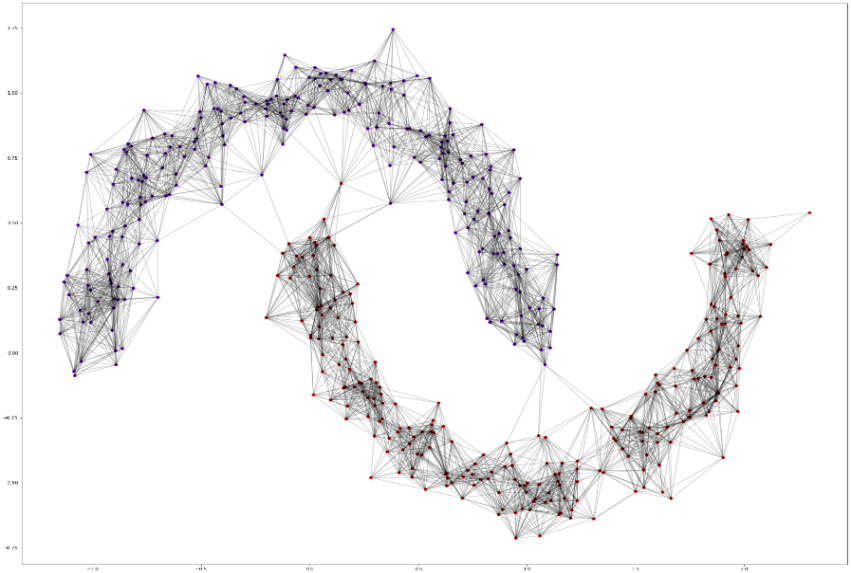}
	\caption{A nearest-neighbors graph of synthetic data.}
	\label{fig:knn}
\end{figure}

The nearest-neighbors algorithm has been widely used in both supervised and unsupervised learning \cite{altman1992introduction,toussaint2005geometric}. The Nearest-Neighbors graph can be constructed on a given set of data $X_{n \times d}$ and a given $K$. This method generates a graph from the records based on their proximity. The strategy for obtaining this graph is presented in Algorithm \ref{alg:knn}. One of the applications of this graph is in subspace spectral clustering, as it can be used as a similarity matrix. An instance of a nearest-neighbors graph of a synthetic dataset is presented in Fig. \ref{fig:knn}. It is noticeable how the graph consists of two weakly-connected sub-graphs which can be broken into as two clusters using spectral clustering.
\begin{algorithm}
	\SetKwInOut{Input}{Input}
	\SetKwInOut{Output}{Output}

	\Input{Dataset $X_{n \times d}$ and integer $K$}
	\Output{Connectivity matrix $M$}
	Initialize $M_{n \times n}$ with zeros\;
	\For{$i\gets1$ \KwTo $n$}
	{
		Initialize $dist_{n \times 1}$\;
		\For{$j\gets1$ \KwTo $n$}
		{
			\eIf{$i \neq j$}
			{
				$dist_j = distance(X_i,X_j)$
			}
			{
				$dist_j = \infty$
			}
		}
		$neighbors \gets$ the indicies of the smallest $K$ elements of $dist$\;
		\For{$neighbor \in neighbors$}
		{
			$M[neighbor,i] = M[i,neighbor] = 1$
		}
	}
	return $M$
	\caption{Nearest-Neighbors Graph}
	\label{alg:knn}
\end{algorithm}

\section{Proposed feature agglomeration and text clustering approach}
\label{sec:method}
In this section, we present the proposed approach, which consists of three main parts: feature extraction using NMF, dimension reduction using LSA and finally, deterministic K-Means clustering. We should note that in this section, the text dataset is already assumed to be tokenized into the term-document matrix $X_{n \times t}$ where $n$ is the number of documents and $t$ is the number of terms. We also assume that this matrix has gone through TF-IDF term weighting. The matrix $X$ is passed along to the method, along with four other parameters which are explained below.
\begin{itemize}
	\item The number of components for NMF, which will serve as the output number of features, $p$.
	\item The number of components for LSA, which can be set to $1$ if LSA is not required, $q$.
	\item The number of neighborhoods for the nearest-neighbors graph, usually set to 5, $r$.
	\item The number of clusters into which the documents are grouped, $K$.
\end{itemize}

\subsection{Feature Extraction using NMF}
\begin{figure}
	\centering
	\includegraphics[width=0.6\textwidth]{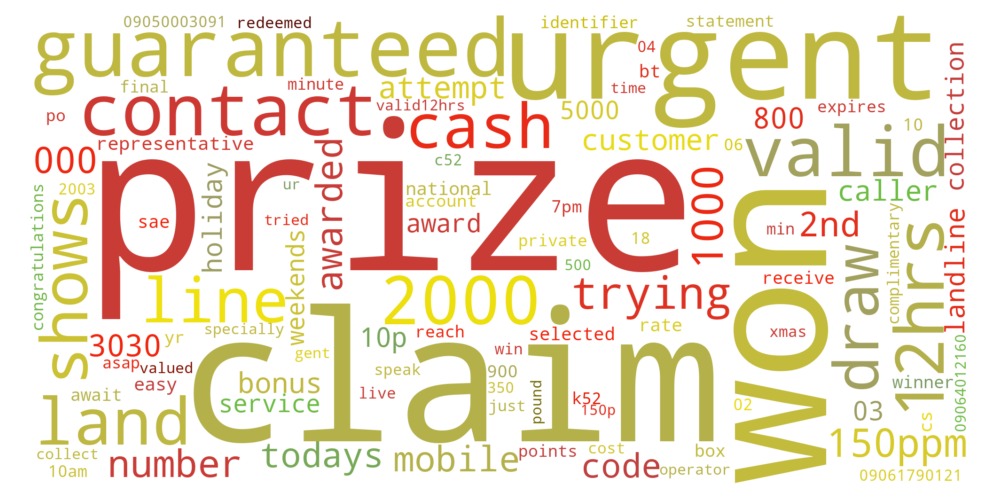}
	\caption{WordCloud of spam terms based on NMF.}
	\label{fig:smsspamcloud}
\end{figure}

\begin{figure}
	\centering
	\begin{tabular}{@{}ccc@{}}
		\subfloat[Economy]{\includegraphics[width=0.32\textwidth]{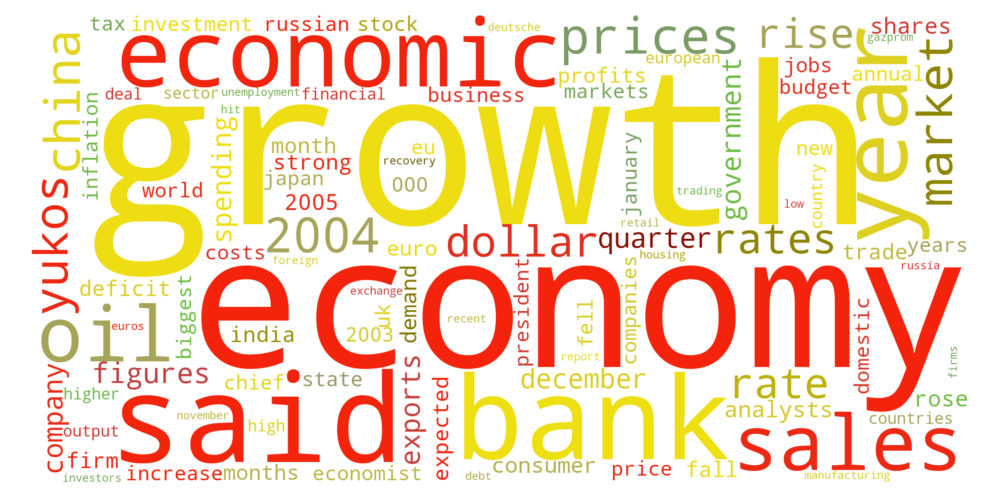}}&
		\subfloat[Entertainment]{\includegraphics[width=0.32\textwidth]{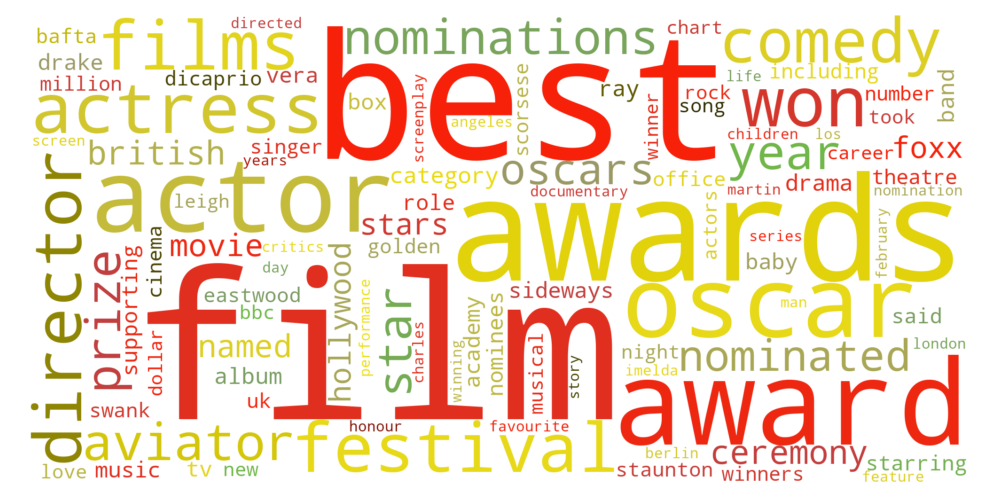}}&
		\subfloat[Politics]{\includegraphics[width=0.32\textwidth]{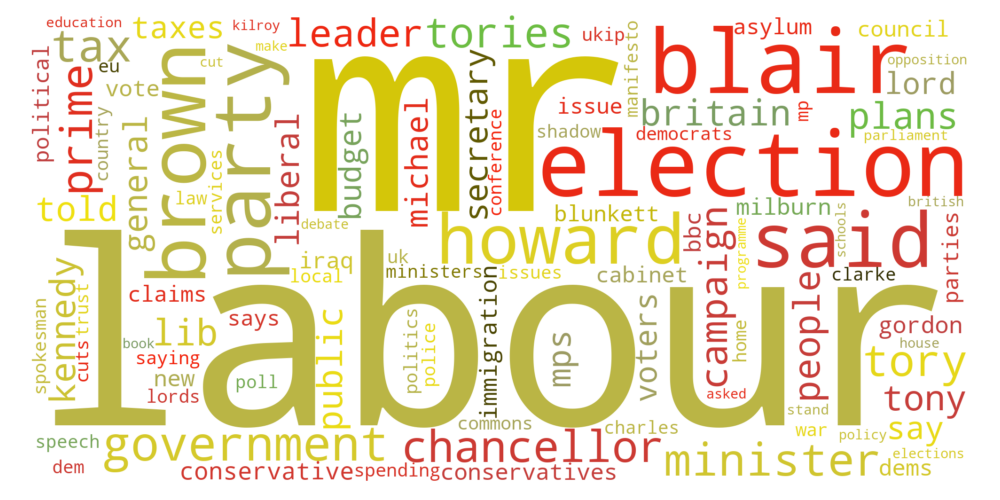}}\\
	\end{tabular}
	\begin{tabular}{@{}cc@{}}
		\subfloat[Sports]{\includegraphics[width=0.33\textwidth]{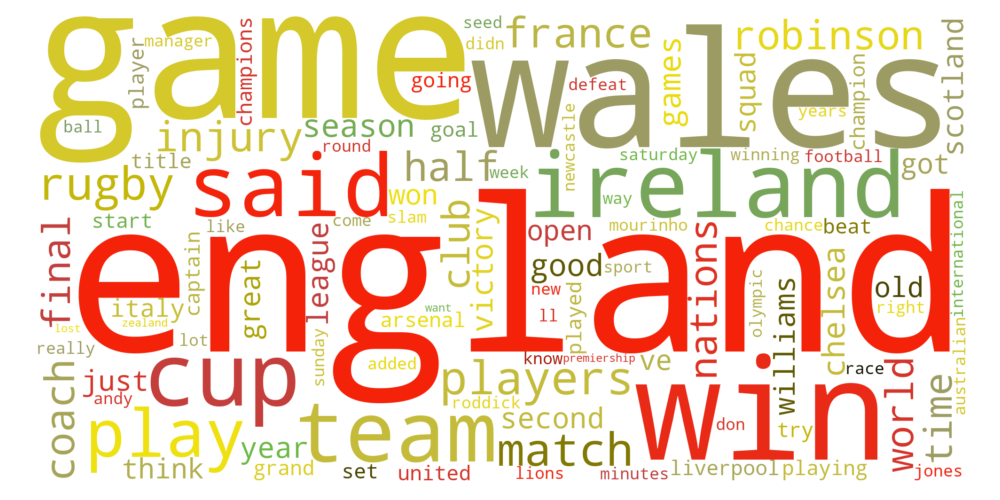}}&
		\subfloat[Technology]{\includegraphics[width=0.33\textwidth]{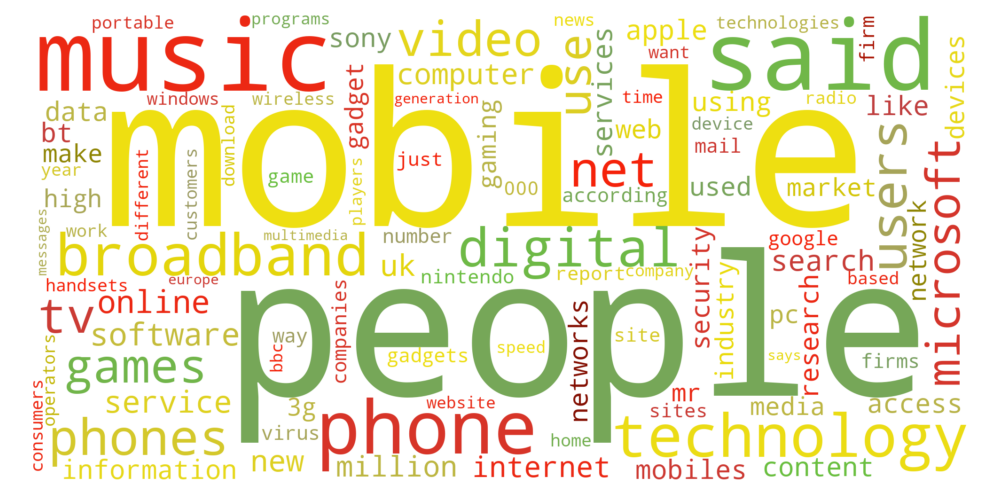}}
	\end{tabular}
	
	\caption{A WordCloud plot of BBC News factorized using NMF.}
	\label{fig:bbcnewswordcloud}
\end{figure}

\begin{figure}
	\centering
	\begin{tabular}{@{}cc@{}}
		\subfloat[Automobiles]{\includegraphics[width=0.33\textwidth]{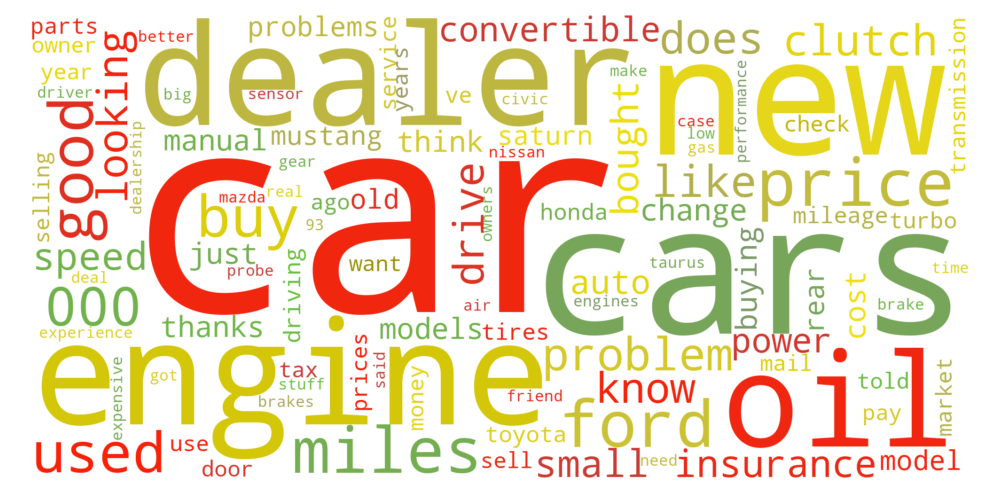}}&
		\subfloat[Motorcycles]{\includegraphics[width=0.33\textwidth]{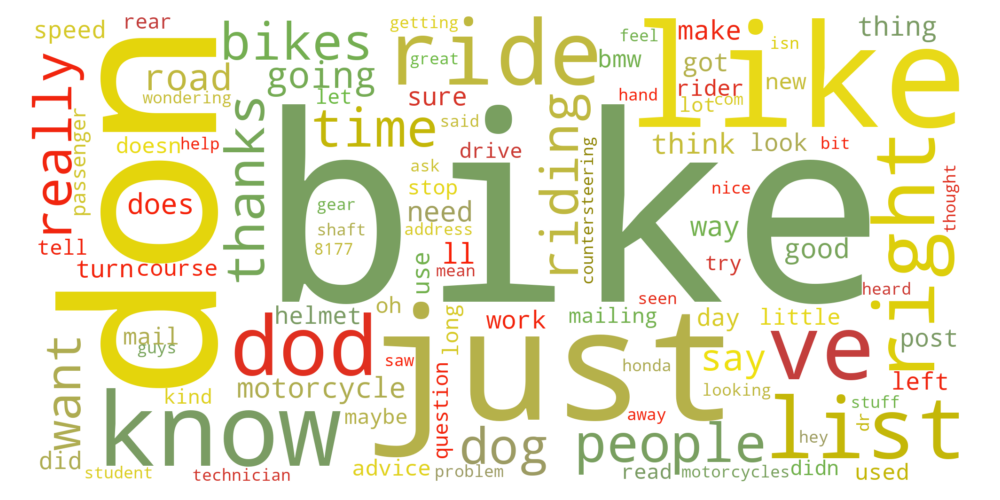}}\\
		\subfloat[Baseball]{\includegraphics[width=0.33\textwidth]{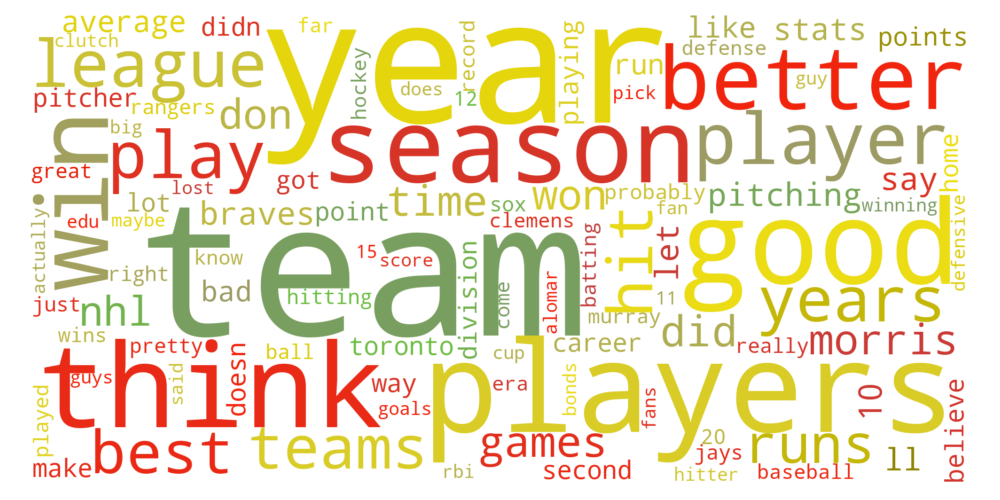}}&
		\subfloat[Hockey]{\includegraphics[width=0.33\textwidth]{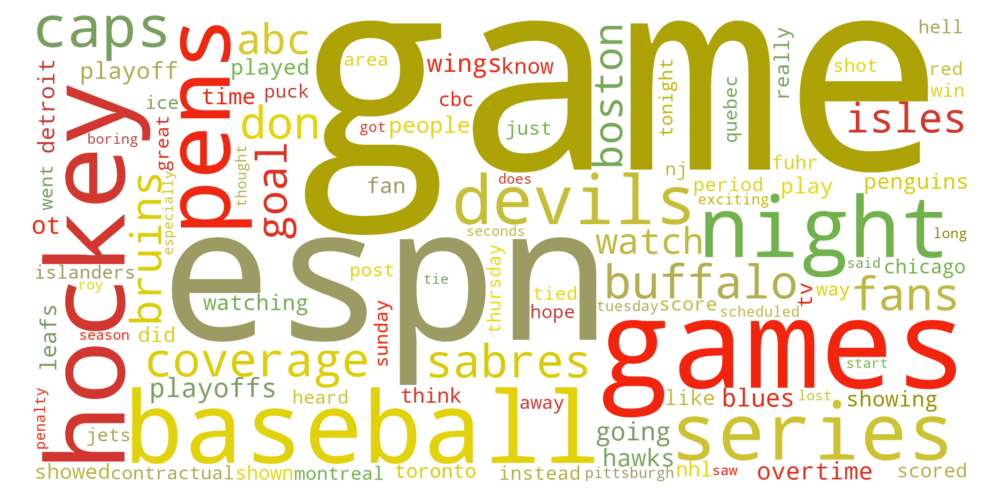}}\\
	\end{tabular}
	
	\caption{A WordCloud plot of 20 Newsgroups - Misc, factorized using NMF.}
	\label{fig:20newswordcloud}
\end{figure}

Nonnegative matrix factorization has been applied in many machine learning problems such as feature extraction and clustering \cite{huang2011enhanced,lu2016semi}. In this paper, we propose a different approach in which NMF helps combine features together in order to create a new and much smaller feature space.
NMF nevertheless requires initialization in order to reach more stability in our case. Because of that, we use singular value decomposition as the initialization method for NMF. Firstly, we start by computing SVD:

\begin{equation}
\label{eq:methodsvd}
X_{n \times t} = U_{ n \times p } \  \Sigma_{p \times p} \  V_{p \times t}^T
\end{equation}
Then, we initialize $W$ and $H$ using Nonnegative Double Singular Value Decomposition method proposed by Boutsidis et al. \cite{boutsidis2008svd}. Afterwards, we enter the NMF optimization phase, which is done using coordinate descent in our experiments.
After the optimization phase, we are left with matrices $W$ and $H$. In our method, we use $H$ in order to group terms together into categories and then represent each category with one feature vector. In other words, we divide terms or the original features into groups and then combine each group's features into one. The number of these groups is the same as the number of components ($p$) selected for NMF, which is our initial parameter. Given the components matrix, $H_{p \times t}$, we define term membership using the following equation:
\begin{flalign}
\label{eq:methodnmfgrouping}
	\begin{aligned}
	&{Term}_i \in {Group}_g, \quad i \in \{ 1,2,…,t \} ,\\
	&\text{where } g = \operatorname*{argmax}_{1 < j < p} {(H_{ij})} ,\\
	&{Term}_i = X_i, \quad \text{where } X_i \text{ is the } i\text{-th column of } X \text{.}
 	\end{aligned}
\end{flalign}
In this section, we will represent each group, which is a set of terms, with a matrix:
\begin{flalign}
\label{eq:methodgroups}
	\begin{aligned}
	&G_g \in M_{n \times \eta_g}, \quad g \in \{ 1,2,...,p \} ,\\
	&\text{where } \eta_g \text{ is the number of terms in group } g \text{.}
	\end{aligned}
\end{flalign}
Followed by that, after the terms are grouped, the new feature space is defined, where the matrix $X^{\prime}$ represents the newly generated space:
\begin{flalign}
\label{eq:methodnewspace}
	\begin{aligned}
	&X^{\prime} = (F_1 F_2 ... F_p)\\
	&\text{where}\\
	&F_g \in M_{n \times 1}, \quad (F_g)_i = \| (G_g)_i \|_2\\
	&g \in \{ 1,2,...,p \}, \quad i \in \{ 1,2,...,n \}\\
	&\text{and } (G_g)_i \text{ is the } i\text{-th row of } G_g \text{.}
	\end{aligned}
\end{flalign}
\begin{figure}[b]
	\centering
	\includegraphics[width=1.0\textwidth]{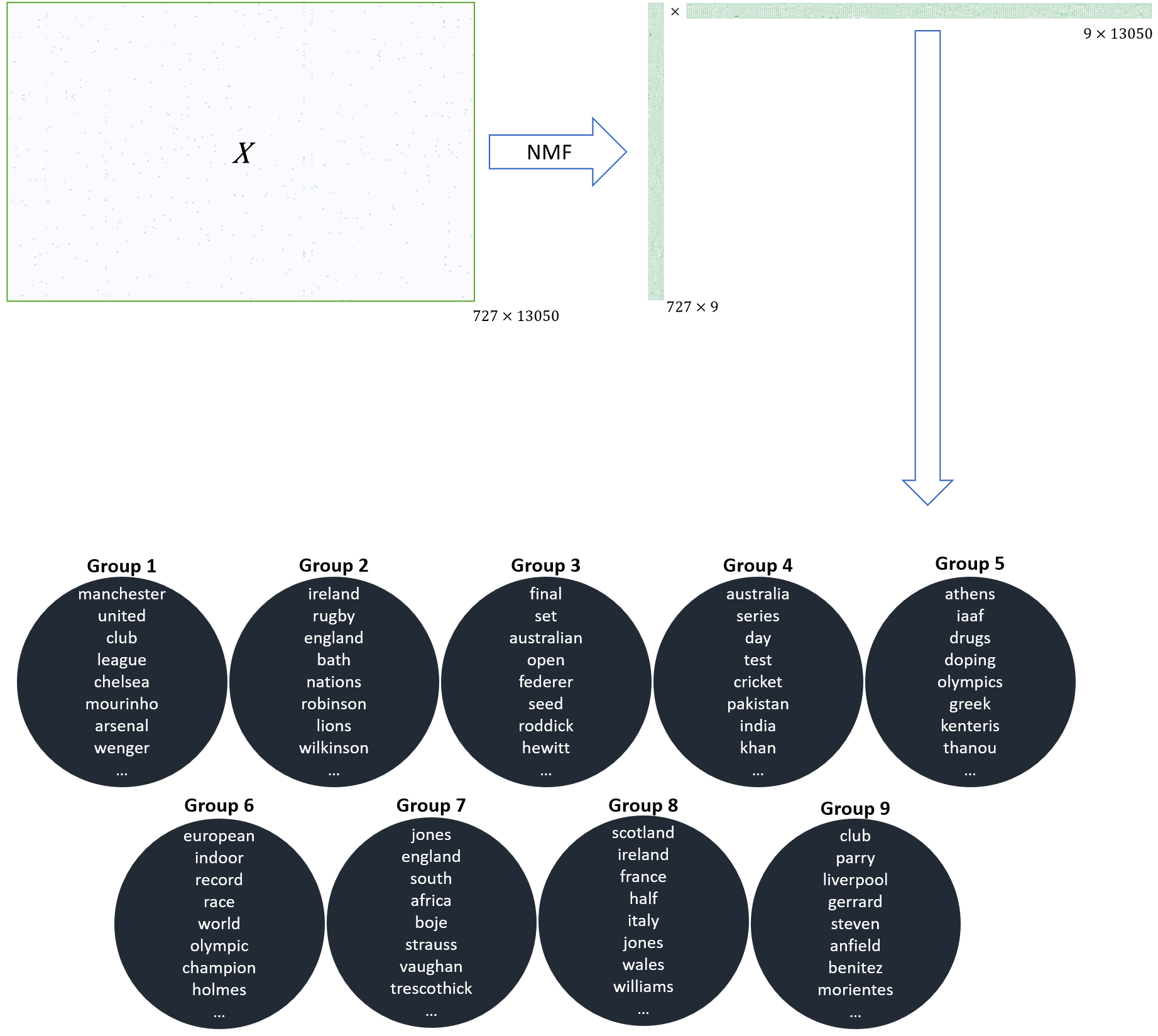}
	\caption{NMF of BBC Sport dataset into 9 components.}
	\label{fig:bbcfactorization}
\end{figure}
The new feature space consists of p feature vectors ($F_g$'s) and each feature is essentially a combination of several terms.
In other words, each group of terms (${Group}_g$), which is a group of $n$-sized vectors, is combined into one $n$-sized vector ($F_g$), which represents a new feature. An example of the BBC Sport \cite{greene2006practical} dataset being factorized into $9$ components ($p = 9$) is presented in Fig. \ref{fig:bbcfactorization}. The terms are sorted by their values in the matrix $H$ (their corresponding column and their assigned row, which corresponds to the maximum value in that column). Furthermore, Fig. \ref{fig:bbcgrouping} presents the groups presented in Fig. \ref{fig:bbcfactorization} matched to the $5$ classes of this dataset (done manually through the observation of the highest-valued terms).
\begin{figure*}
	\centering
	\begin{tabular}{@{}ccc@{}}
		\subfloat[Football]{\includegraphics[width=0.33\textwidth]{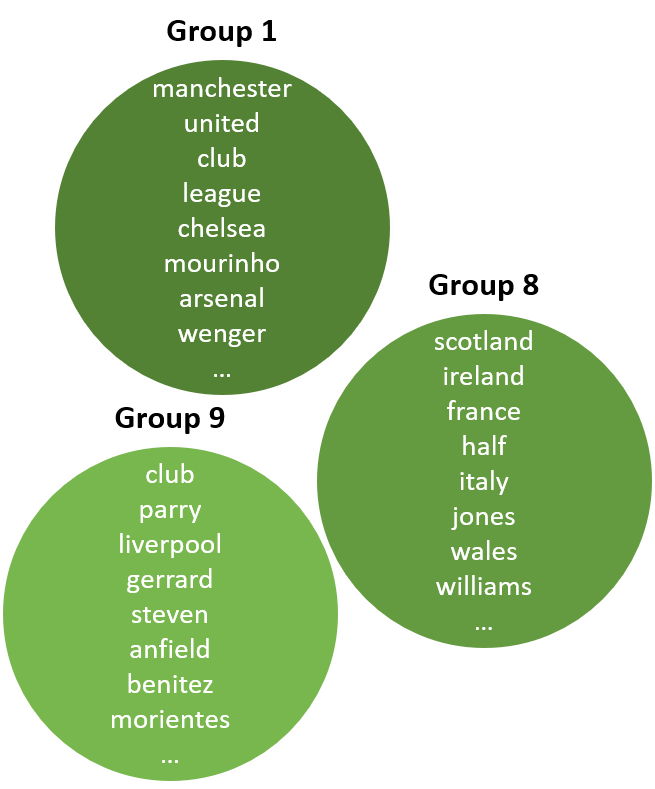}}&
		\subfloat[Rugby]{\includegraphics[width=0.33\textwidth]{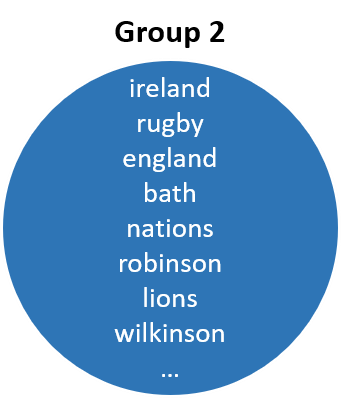}}&
		\subfloat[Tennis]{\includegraphics[width=0.33\textwidth]{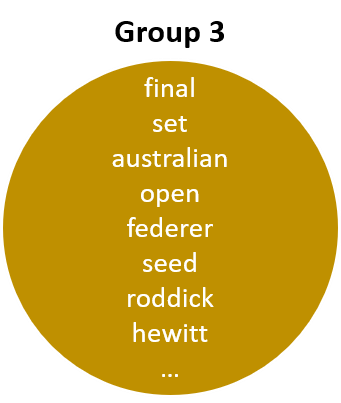}}\\
	\end{tabular}
	\begin{tabular}{@{}cc@{}}
		\subfloat[Athletic]{\includegraphics[width=0.5\textwidth]{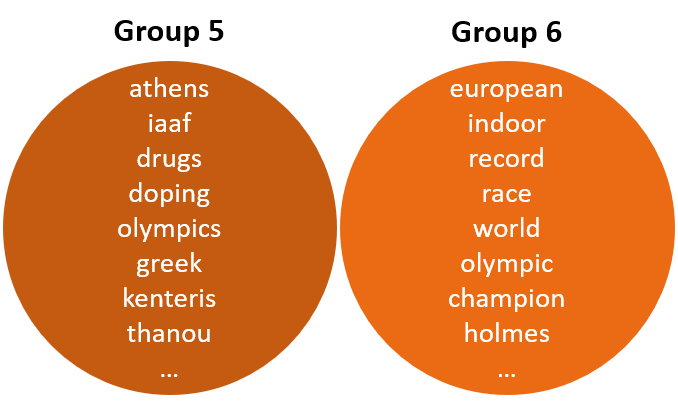}}&
		\subfloat[Cricket]{\includegraphics[width=0.5\textwidth]{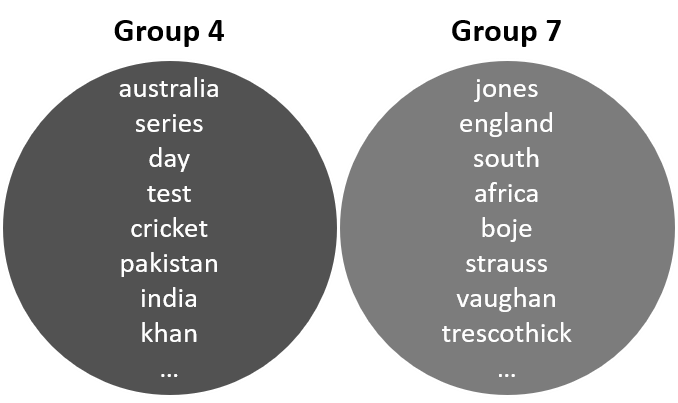}}\\
	\end{tabular}
	
	\caption{The groups in Fig. \ref{fig:bbcfactorization} matched to their closest matching classes.}
	\label{fig:bbcgrouping}
\end{figure*}
A WordCloud plot of possible spam words extracted using NMF is presented in Fig. \ref{fig:smsspamcloud}. Moreover, WordCloud plots of BBC News \cite{greene2006practical} and 20 Newsgroups (Miscellaneous) \cite{lang1995newsweeder} are presented in figures \ref{fig:bbcnewswordcloud} and \ref{fig:20newswordcloud}. These plots were generated using the Python library Word Cloud \cite{wordcloudlib}. In each plot, the top 100 terms from each NMF group are shown, and the size of each term represents its relative frequency obtained from the matrix $H$ from NMF.
It should be noted that the colors are randomly generated and point towards no specific detail. These plots help visualize how NMF effectively separates the terms into groups.

\subsection{Latent Semantic Analysis Reduction}
This phase is optional in the proposed approach, as it may be useful in specific cases in which it can boost clustering performance by projecting the data into a different space. As explained in the previous section, LSA uses singular value decomposition in order to reduce dimensionality. If the parameter $q$ is set to $1$, no LSA reduction is performed and this section is bypassed. If not, a new feature subspace is defined:
\begin{flalign}
\label{eq:methodlsa}
	\begin{aligned}
	&X^{\prime \prime} = X^{\prime}, \quad q = 1,\\
	&X^{\prime \prime} = LSA(X^{\prime}, q) , \quad \text{otherwise.}
	\end{aligned}
\end{flalign}
This step is made optional, due to the fact that LSA can sometimes result in more convenient results when applied to the newly generated feature space.
\subsection{Clustering using KNN initialization}
\begin{figure*}
	\centering
	\includegraphics[height=\textheight,keepaspectratio]{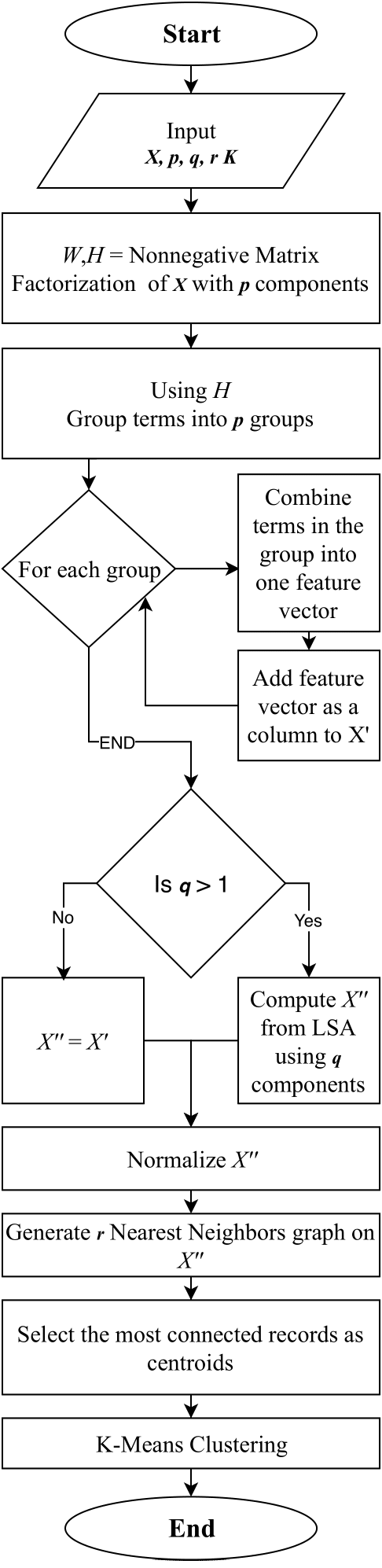}
	\caption{Flowchart of the proposed method.}
	\label{fig:flowchart}
\end{figure*}

After the dimensionality reduction using NMF and LSA, the proposed method tries to separate the documents into clusters. The proposed method clusters data using spherical K-Means, which normalizes the data through dividing each record by its norm. This essentially maps the records onto the $p$-dimensional hyper-sphere ($q$-dimensional if LSA is also applied). This results in spherical clustering which is widely employed in text mining. Afterwards, we employ a new centroid initialization strategy for K-Means which is deterministic in nature.
It is a widely-known fact that K-Means is highly affected by its initialization, which is why K-Means++ initialization is used. This initialization is far more effective, but is nevertheless also stochastic in nature. Therefore, we propose a nearest-neighbors-based centroid initialization for K-Means.
\begin{figure}
	\centering
	\includegraphics[width=0.75\textwidth]{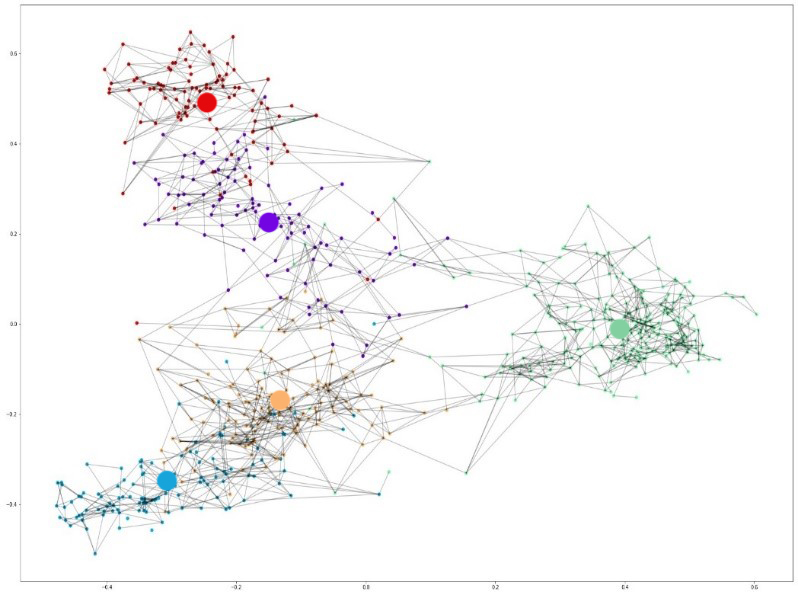}
	\caption{Final centroids after a K-Means on the dataset BBC Sport initialized by the proposed method.}
	\label{fig:kmeansinit}
\end{figure}
The new strategy works simply by creating an $r$-nearest-neighbors graph of the records, and taking the top $K$ nodes (records) with the most connections. Since the nearest-neighbors algorithm connects each record to its $r$ nearest neighbors, any record in a highly dense area could have many connections and therefore be suitable for being chosen as the centroid. After these records are chosen, they are passed along to K-Means as the initial centroids and K-Means clusters the data in a very deterministic way.
If we consider the results from NMF and LSA to be stable and almost deterministic (which is usually the case), this results in a standard deviation of $0$ in our results. An example of the final centroids initialized by this strategy is presented in Fig. \ref{fig:kmeansinit}. In this figure, the dataset has been reduced to $2$-dimensional space using Principal Component Analysis, in order to help with its visualization.
An overview of the proposed approach is also presented in Fig. \ref{fig:flowchart}.

\section{Experiment details}
\label{sec:experiments}
We conducted experiments on several datasets in order to compare the proposed approach to some of the recently proposed approaches, as well as some of the most practically used methods. The metrics used for comparison are the clustering accuracy (purity), Normalized Mutual Information (NMI), and Adjusted Rand Index (ARI). The datasets used in our experiments are provided in Table \ref{tab:datasets}.

\begin{table*}[!h]
	\centering
	\caption{An overview of the datasets used in our experiments}
	\label{tab:datasets}
	\begin{tabular}{p{0.34\textwidth}p{0.12\textwidth}p{0.1\textwidth}p{0.07\textwidth}p{0.07\textwidth}p{0.09\textwidth}}
		\hline\noalign{\smallskip}
		Dataset&Abbreviation&Documents&Terms&Classes& Reference\\
		\noalign{\smallskip}\hline\noalign{\smallskip}
		20 Newsgroups – Computer&20COMP&4582&64139&5&\cite{lang1995newsweeder}\\
		20 Newsgroups – Politics&20POL&2287&41254&3&\cite{lang1995newsweeder}\\
		20 Newsgroups – Miscellaneous&20MISC&3648&39836&4&\cite{lang1995newsweeder}\\
		20 Newsgroups – Religion&20REL&2195&40149&3&\cite{lang1995newsweeder}\\
		20 Newsgroups – Science&20SCI&3617&54373&4&\cite{lang1995newsweeder}\\
		AG News&AG&1180&7054&4&\cite{gulli2004}\\
		BBC Sport&BBCSP&727&13050&5&\cite{greene2006practical}\\
		BBC News&BBC&2127&29392&5&\cite{greene2006practical}\\
		DMOZ&DMOZ&3870&8758&13&\cite{dmozdataset}\\
		SMS&SMS&1509&4626&2&\cite{almeida2011,almeida2013towards}\\
		WebKB&WEB&1773&27537&7&\cite{cmu1997www}\\
		WebAce&WEBACE&1151&6807&21&\cite{han1998webace}\\
		\noalign{\smallskip}\hline
	\end{tabular}
\end{table*}

We should note that these datasets were under-sampled in order to balance the dataset and reduce computational complexity.
20 Newsgroups dataset \cite{lang1995newsweeder} is a text document set containing over 18,000 documents in 20 groups. The groups can be divided into 5 categories, where each category has different classes. In order to separate results and reduce computation, we divided the dataset into the 5 categories, considering each separately.
AG News \cite{gulli2004} on the other hand is a corpus of news articles collected from over 2,000 webpages.
BBC dataset \cite{greene2006practical} consists of two sets, one which categorizes BBC News transcripts into 5 categories: Business, Entertainment, Politics, Sport, Technology, and the other categorizes BBC Sports transcripts into 5 groups: Football, Rugby, Tennis, Athletic, and Cricket.
DMOZ \cite{dmozdataset} is a dataset containing text information from the DMOZ (Open Dictionary Project) which contains 13 classes.
SMS \cite{almeida2013towards} dataset on the other hand contains 4,827 non-spam and 747 spam text messages. Due to the high imbalance between the two classes, we resampled the data into almost 1,500 documents containing almost an equal percentage of each class.
WebKB \cite{cmu1997www} contains text information collected from webpages and is categorized into 7 classes: Staff, Department, Project, Course, Faculty, Student and Other.
WebAce \cite{han1998webace} dataset consists of over 1,500 web documents in 21 categories, which has also been samples.
We conducted our experiments on a Windows personal computer with 16 GB of RAM (swap space has been used by the OS) and a quad core Intel Core i7 processor clocked between 2.6 to 3.4 GHz. Our implementations were done in Python and we used the following libraries: Numpy \cite{van2011numpy}, Scikit-Learn \cite{pedregosa2011scikit} and Imbalanced Learn \cite{lemaitre2017imbalanced}.

\section{Experiment results and discussion}
\label{sec:discussion}
In this section, we present the results of our experiments. We divide our results into three subsections. In the first, we compare the proposed method (NMF-FR) to K-Means (KM), Spherical K-Means (SKM), Genetic Algorithm and K-Means Clustering (GAKM) \cite{garg2018performance}, and Spectral Clustering with Particle Swarm Optimization (SCPSO) \cite{janani2019text}. In the second section, we compare the proposed approach to one of the most widely used dimension reduction and sentiment analysis methods, LSA, clustered using K-Means (LSAKM).
In our experiments, K-Means was initialized using K-Means++. Finally, we evaluate the stability of the proposed approach using the standard deviation from the clustering results.
\subsection{Comparison to clustering methods}
We present the results of the experiments in Tables \ref{tab:cluster_acc} - \ref{tab:cluster_ari}.

\begin{table}[h!]
	\centering
	\caption{Accuracy of the proposed method compared to clustering methods}
	\label{tab:cluster_acc}
	\begin{tabular}{p{0.15\textwidth}p{0.1\textwidth}p{0.1\textwidth}p{0.1\textwidth}p{0.1\textwidth}p{0.1\textwidth}}
		\hline\noalign{\smallskip}
		Dataset&GAKM&KM&SCPSO&SKM&NMF-FR\\
		\noalign{\smallskip}\hline\noalign{\smallskip}
		
		20COMP&0.411261&0.396333&0.216761&0.469708&\textbf{0.544304}\\ 
		20POL&0.49777&0.53415&0.553126&0.602973&\textbf{0.613467}\\ 
		20MISC&0.370504&0.522807&0.26409&0.532072&\textbf{0.772204}\\ 
		20REL&0.471891&0.450843&0.550524&0.477267&\textbf{0.550797}\\ 
		20SCI&0.557368&0.552391&0.600553&0.700581&\textbf{0.764999}\\ 
		AG&0.374237&0.438305&0.490847&0.451186&\textbf{0.674576}\\ 
		BBCSP&0.744154&0.832187&0.928748&0.826685&\textbf{0.965612}\\ 
		BBC&0.80771&0.807804&0.640903&0.877762&\textbf{0.921016}\\ 
		DMOZ&0.352817&0.373747&0.468941&0.518346&\textbf{0.543411}\\ 
		SMS&0.67104&0.675944&0.57283&0.746587&\textbf{0.809145}\\ 
		WEB&0.41771&0.418951&0.323181&0.439368&\textbf{0.491258}\\ 
		WEBACE&0.388532&0.380712&0.372893&0.385578&\textbf{0.448306}\\
		
		\noalign{\smallskip}\hline
	\end{tabular}
\end{table}
\begin{table}[h!]
	\centering
	\caption{NMI Comparison between the proposed method and clustering methods}
	\label{tab:cluster_nmi}
	\begin{tabular}{p{0.15\textwidth}p{0.1\textwidth}p{0.1\textwidth}p{0.1\textwidth}p{0.1\textwidth}p{0.1\textwidth}}
		\hline\noalign{\smallskip}
		Dataset&GAKM&KM&SCPSO&SKM&NMF-FR\\
		\noalign{\smallskip}\hline\noalign{\smallskip}
		
		20COMP&0.1757&0.151818&0.001881&0.211848&\textbf{0.239178}\\ 
		20POL&0.124027&0.224761&0.198366&0.287887&\textbf{0.336881}\\ 
		20MISC&0.153722&0.280174&0.001704&0.316807&\textbf{0.482384}\\ 
		20REL&0.045574&0.052925&\textbf{0.11356}&0.052902&0.096347\\ 
		20SCI&0.298135&0.309364&0.385496&0.436442&\textbf{0.476141}\\
		AG&0.136012&0.181492&0.259041&0.174582&\textbf{0.314543}\\ 
		BBCSP&0.654929&0.732905&0.857578&0.781206&\textbf{0.891782}\\ 
		BBC&0.709906&0.707077&0.537028&0.785033&\textbf{0.784895}\\ 
		DMOZ&0.329851&0.326525&0.432909&\textbf{0.436235}&0.419595\\ 
		SMS&0.165144&0.1769&0.021677&0.219879&\textbf{0.304682}\\ 
		WEB&0.248054&0.254047&0.124423&0.264204&\textbf{0.286988}\\ 
		WEBACE&0.374864&0.374677&0.367674&0.381486&\textbf{0.403161}\\ 
		
		\noalign{\smallskip}\hline
	\end{tabular}
\end{table}
\begin{table}[h!]
	\centering
	\caption{ARI Comparison between the proposed method and clustering methods}
	\label{tab:cluster_ari}
	\begin{tabular}{p{0.15\textwidth}p{0.1\textwidth}p{0.1\textwidth}p{0.1\textwidth}p{0.1\textwidth}p{0.1\textwidth}}
		\hline\noalign{\smallskip}
		Dataset&GAKM&KM&SCPSO&SKM&NMF-FR\\
		\noalign{\smallskip}\hline\noalign{\smallskip}
		
		20COMP&0.090171&0.075304&0.000097&0.150885&\textbf{0.20118}\\ 
		20POL&0.085958&0.100406&0.151&0.208516&\textbf{0.241515}\\ 
		20MISC&0.070536&0.18378&5.29E-05&0.238066&\textbf{0.497897}\\ 
		20REL&0.028029&0.022696&\textbf{0.131546}&0.042575&0.086511\\ 
		20SCI&0.16944&0.180342&0.286651&0.35856&\textbf{0.47049}\\ 
		AG&0.034796&0.082022&0.168704&0.114249&\textbf{0.331825}\\ 
		BBCSP&0.485238&0.673881&0.83687&0.682754&\textbf{0.909467}\\ 
		BBC&0.646132&0.636038&0.41667&0.769899&\textbf{0.820115}\\ 
		DMOZ&0.070285&0.086746&0.19843&0.270969&\textbf{0.326107}\\ 
		SMS&0.112427&0.145561&0.005877&0.248991&\textbf{0.381879}\\ 
		WEB&0.152072&0.145514&0.077537&0.178062&\textbf{0.221728}\\ 
		WEBACE&0.169217&0.165472&0.149198&0.17589&\textbf{0.198501}\\
		
		\noalign{\smallskip}\hline
	\end{tabular}
\end{table}
The results indicate that proposed method exceeds the other clustering approaches in all the metrics in almost every dataset with notable increase in results. Nevertheless, the true advantage of the proposed approach is its stability, which will be further discussed later. Another notable advantage of the proposed method is the dimension reduction which can decrease computational complexity, especially when working with term-document matrices in which the number of terms (or features) is much greater than the number of documents (or records).

\subsection{Comparison to LSA}
We present the results of the experiments comparing LSA combined with Spherical K-Means++ (LSAKM) and the proposed approach (NMF-FR) in Tables \ref{tab:lsa_acc} - \ref{tab:lsa_ari}. We should note that in LSAKM, data normalization is conducted as well before K-Means clustering. LSAKM also uses K-Means++ for clustering.

\begin{figure}[h!]
	\centering
	\begin{tikzpicture}
	\begin{axis}[
	width=0.99\textwidth,
	height=0.38\textheight,
	grid=both,
	xlabel={Dataset},
	ylabel={Accuracy},
	tick label style={font=\tiny},
	ymin=0.3, ymax=1,
	xtick=data,
	x tick label style={rotate=45,anchor=east},
	symbolic x coords={20COMP,20POL,20MISC,20REL,20SCI,AG,BBCSP,BBC,DMOZ,SMS,WEB,WEBACE},
	point meta=explicit symbolic,
	x label style={at={(axis description cs:0.5,-0.05)},anchor=north},
	]
	
	\addplot[
	color=darkgray,
	mark=diamond,
	ultra thick,
	mark size=3pt,
	]
	coordinates {
		(20COMP,0.471497)(20POL,0.540096)(20MISC,0.705647)(20REL,0.557084)(20SCI,0.751728)(AG,0.580339)(BBCSP,0.941403)(BBC,0.926093)(DMOZ,0.507545)(SMS,0.794964)(WEB,0.461365)(WEBACE,0.463076)
	};
	\addlegendentry{LSAKM}

	\addplot[
	color=green,
	mark=square,
	ultra thick,
	mark size=3pt,
	]
	coordinates {
		(20COMP,0.544304)(20POL,0.613467)(20MISC,0.772204)(20REL,0.550797)(20SCI,0.764999)(AG,0.674576)(BBCSP,0.965612)(BBC,0.921016)(DMOZ,0.543411)(SMS,0.809145)(WEB,0.491258)(WEBACE,0.448306)
	};
	\addlegendentry{NMF-FR}

	\end{axis}
	\end{tikzpicture}
	\caption{LSAKM accuracy compared to the proposed method.}
	\label{fig:lsavsnmffr}
\end{figure}
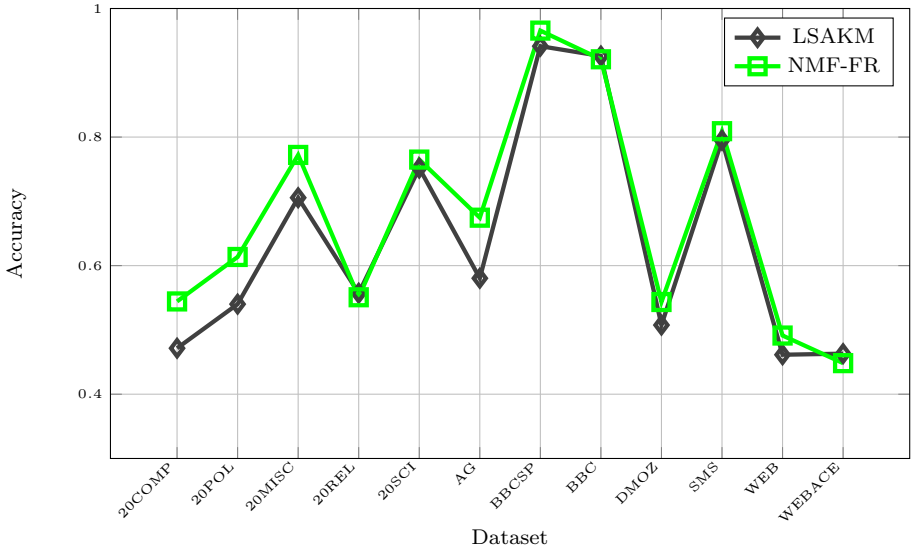

\begin{table}[!h]
	\centering
	\caption*{Comparison between LSA-KMeans and the proposed method}
	\begin{minipage}{.45\textwidth}
		\centering
		\caption{Accuracy}
		\begin{tabular}{p{0.2\textwidth}p{0.25\textwidth}p{0.25\textwidth}}
			\hline\noalign{\smallskip}
			Dataset&LSAKM&NMF-FR\\
			\noalign{\smallskip}\hline\noalign{\smallskip}
			
			20COMP&0.471497&\textbf{0.544304}\\ 
			20POL&0.540096&\textbf{0.613467}\\ 
			20MISC&0.705647&\textbf{0.772204}\\ 
			20REL&\textbf{0.557084}&0.550797\\ 
			20SCI&0.751728&\textbf{0.764999}\\ 
			AG&0.580339&\textbf{0.674576}\\ 
			BBCSP&0.941403&\textbf{0.965612}\\ 
			BBC&\textbf{0.926093}&0.921016\\ 
			DMOZ&0.507545&\textbf{0.543411}\\ 
			SMS&0.794964&\textbf{0.809145}\\ 
			WEB&0.461365&\textbf{0.491258}\\ 
			WEBACE&\textbf{0.463076}&0.448306\\ 
			
			\noalign{\smallskip}\hline
		\end{tabular}
		\label{tab:lsa_acc}
	\end{minipage}
	\begin{minipage}{.45\textwidth}
		\centering
		\caption{NMI}
		\begin{tabular}{p{0.2\textwidth}p{0.25\textwidth}p{0.25\textwidth}}
			\hline\noalign{\smallskip}
			Dataset&LSAKM&NMF-FR\\
			\noalign{\smallskip}\hline\noalign{\smallskip}
			
			20COMP&0.187843&\textbf{0.239178}\\ 
			20POL&0.168456&\textbf{0.336881}\\ 
			20MISC&0.448408&\textbf{0.482384}\\ 
			20REL&\textbf{0.102754}&0.096347\\ 
			20SCI&0.456323&\textbf{0.476141}\\ 
			AG&0.267823&\textbf{0.314543}\\ 
			BBCSP&0.864227&\textbf{0.891782}\\ 
			BBC&\textbf{0.807666}&0.784895\\ 
			DMOZ&0.395593&\textbf{0.419595}\\ 
			SMS&\textbf{0.315545}&0.304682\\ 
			WEB&0.273446&\textbf{0.286988}\\ 
			WEBACE&\textbf{0.44652}&0.403161\\ 
			
			\noalign{\smallskip}\hline
		\end{tabular}
		\label{tab:lsa_nmi}
	\end{minipage}
\end{table}
\begin{table}[h!]
	\centering
	\caption{Comparison between LSA-KMeans and the proposed method in ARI}
	\label{tab:lsa_ari}
	\begin{tabular}{p{0.2\textwidth}p{0.25\textwidth}p{0.25\textwidth}}
		\hline\noalign{\smallskip}
		Dataset&LSAKM&NMF-FR\\
		\noalign{\smallskip}\hline\noalign{\smallskip}
		
		20COMP&0.146904&\textbf{0.20118}\\ 
		20POL&0.129691&\textbf{0.241515}\\ 
		20MISC&0.446994&\textbf{0.497897}\\ 
		20REL&\textbf{0.106373}&0.086511\\ 
		20SCI&0.463747&\textbf{0.47049}\\ 
		AG&0.228503&\textbf{0.331825}\\ 
		BBCSP&0.873331&\textbf{0.909467}\\ 
		BBC&\textbf{0.836696}&0.820115\\ 
		DMOZ&0.289523&\textbf{0.326107}\\ 
		SMS&0.347382&\textbf{0.381879}\\ 
		WEB&0.192322&\textbf{0.221728}\\ 
		WEBACE&\textbf{0.228777}&0.198501\\
		
		\noalign{\smallskip}\hline
	\end{tabular}
\end{table}
As it can be observed, the proposed method shows considerable, if not significant improvement over simple LSA + K-Means. Furthermore, a chart plotting the accuracies of LSAKM and the proposed method is presented in Fig. \ref{fig:lsavsnmffr}. LSA in practice relies on a singular value decomposition of the term-document matrix, which can lead to different results each time. Followed by that, K-Means++ itself is a stochastic measure. These two issues question the stability of this method's results. The proposed method on the other hand does not rely on the exact output from NMF, as it discretizes its results. Meanwhile, the clustering initialization is done using a deterministic method in the proposed approach, which makes it far more stable than LSAKM.

\begin{figure}[!h]
	\centering
	\begin{tabular}{@{}cc@{}}
		\subfloat[Original Space]{\includegraphics[width=0.475\textwidth]{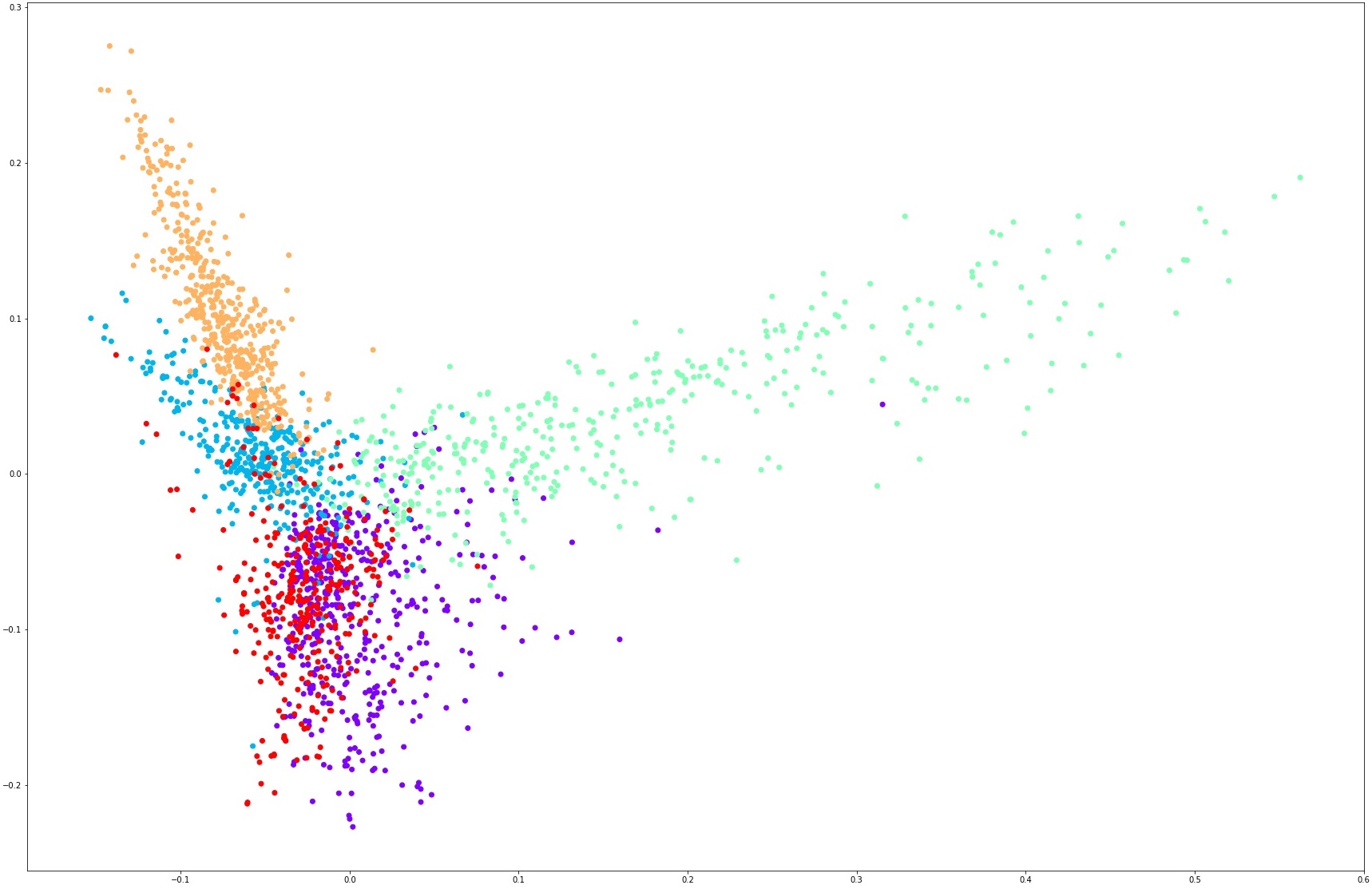}}&
		\subfloat[LSA Space]{\includegraphics[width=0.475\textwidth]{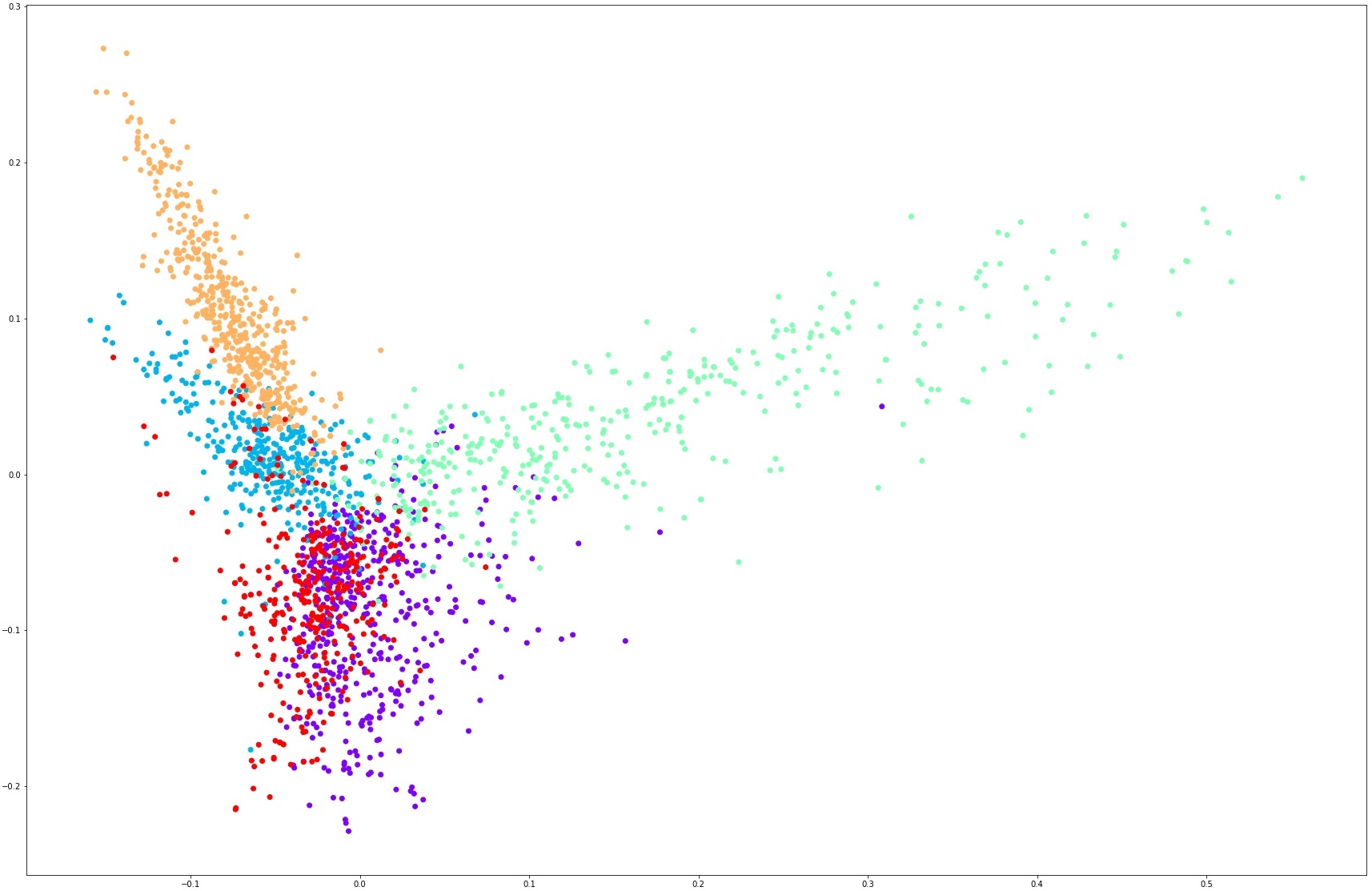}}
		\\
	\end{tabular}
	\begin{tabular}{@{}c@{}}
		\subfloat[NMF-FR Space]{\includegraphics[width=0.85\textwidth]{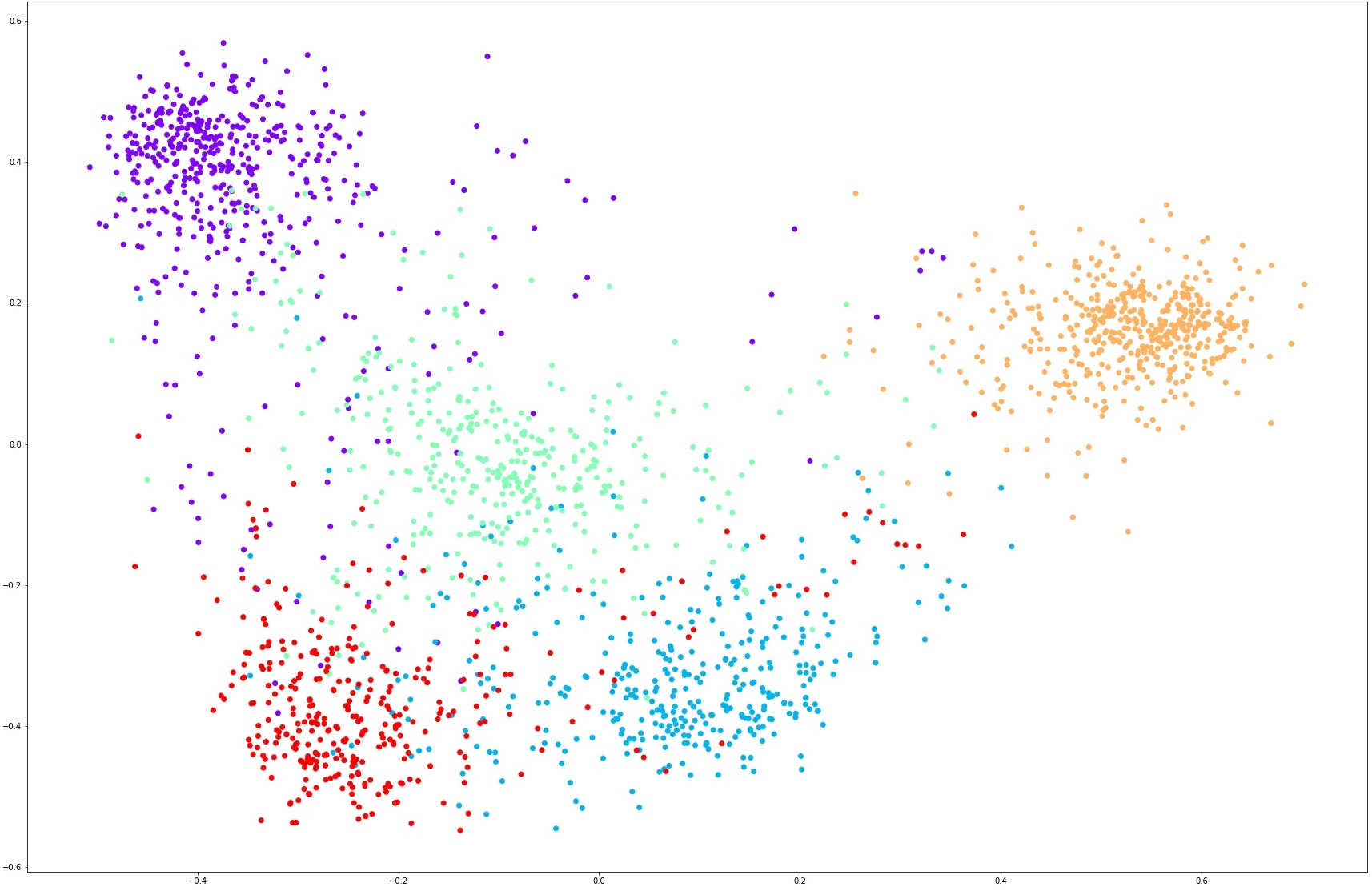}}
	\end{tabular}
	
	\caption{The original data space (a), the space generated after LSA (b), and the space generated after the proposed method was applied (c) to BBC News.}
	\label{fig:pcacomparison1}
\end{figure}

\begin{figure*}
	\centering
	\begin{tabular}{@{}cc@{}}
		\subfloat[Ground Truth]{\includegraphics[width=0.45\textwidth]{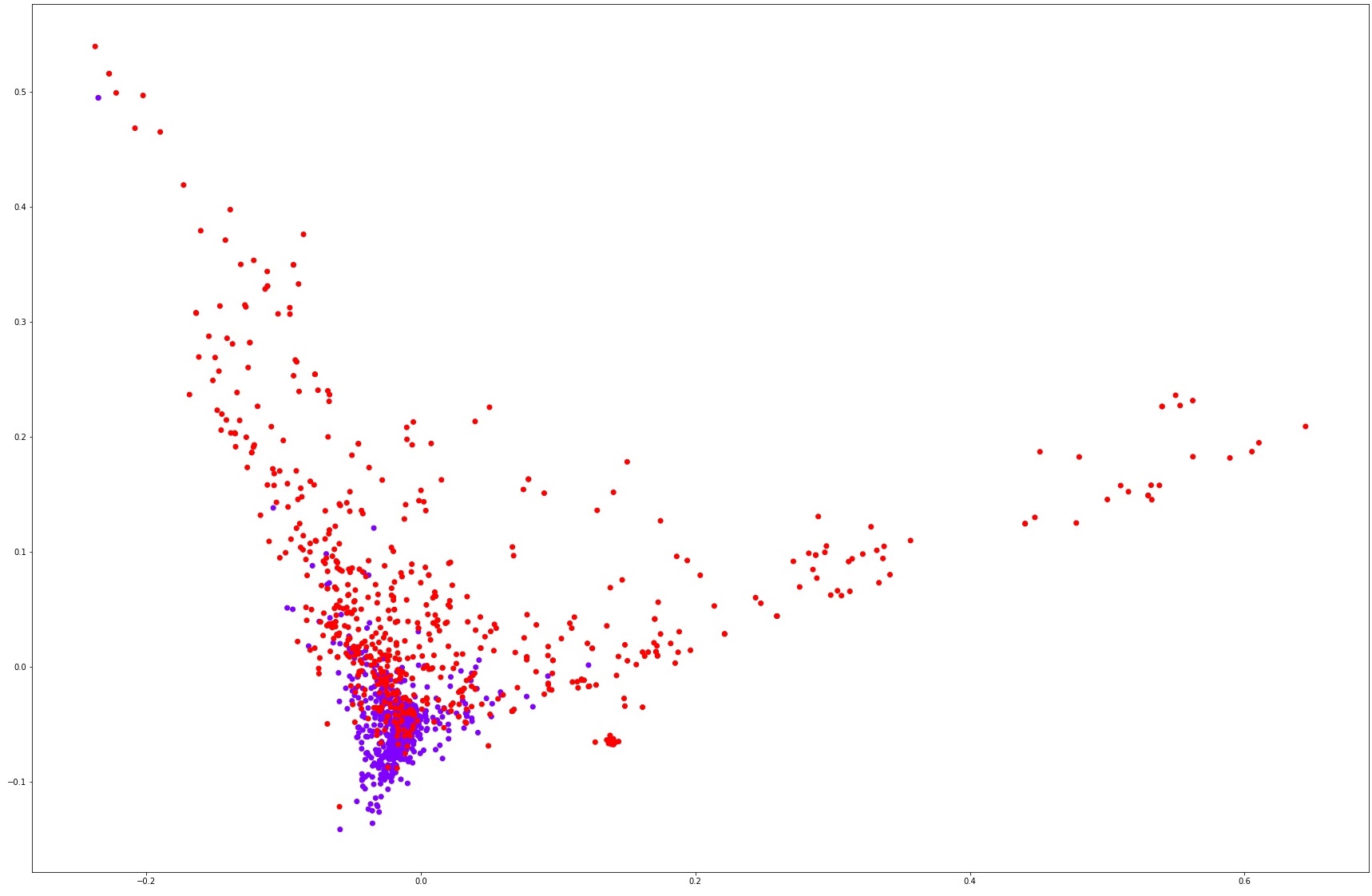}}&
		\subfloat[Prediction]{\includegraphics[width=0.45\textwidth]{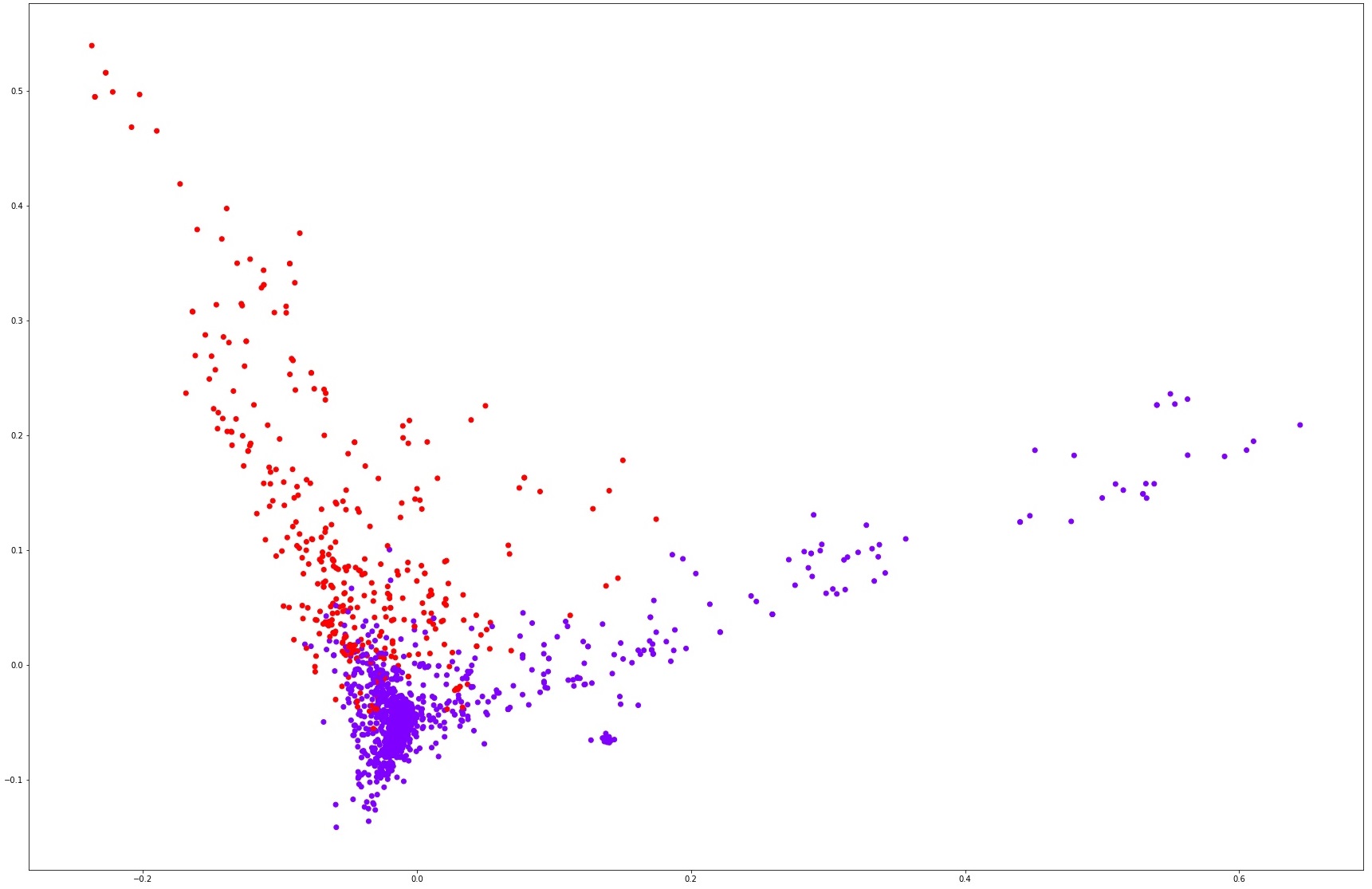}}
		\\
	\end{tabular}
	\begin{tabular}{@{}cc@{}}
		\subfloat[Ground Truth]{\includegraphics[width=0.45\textwidth]{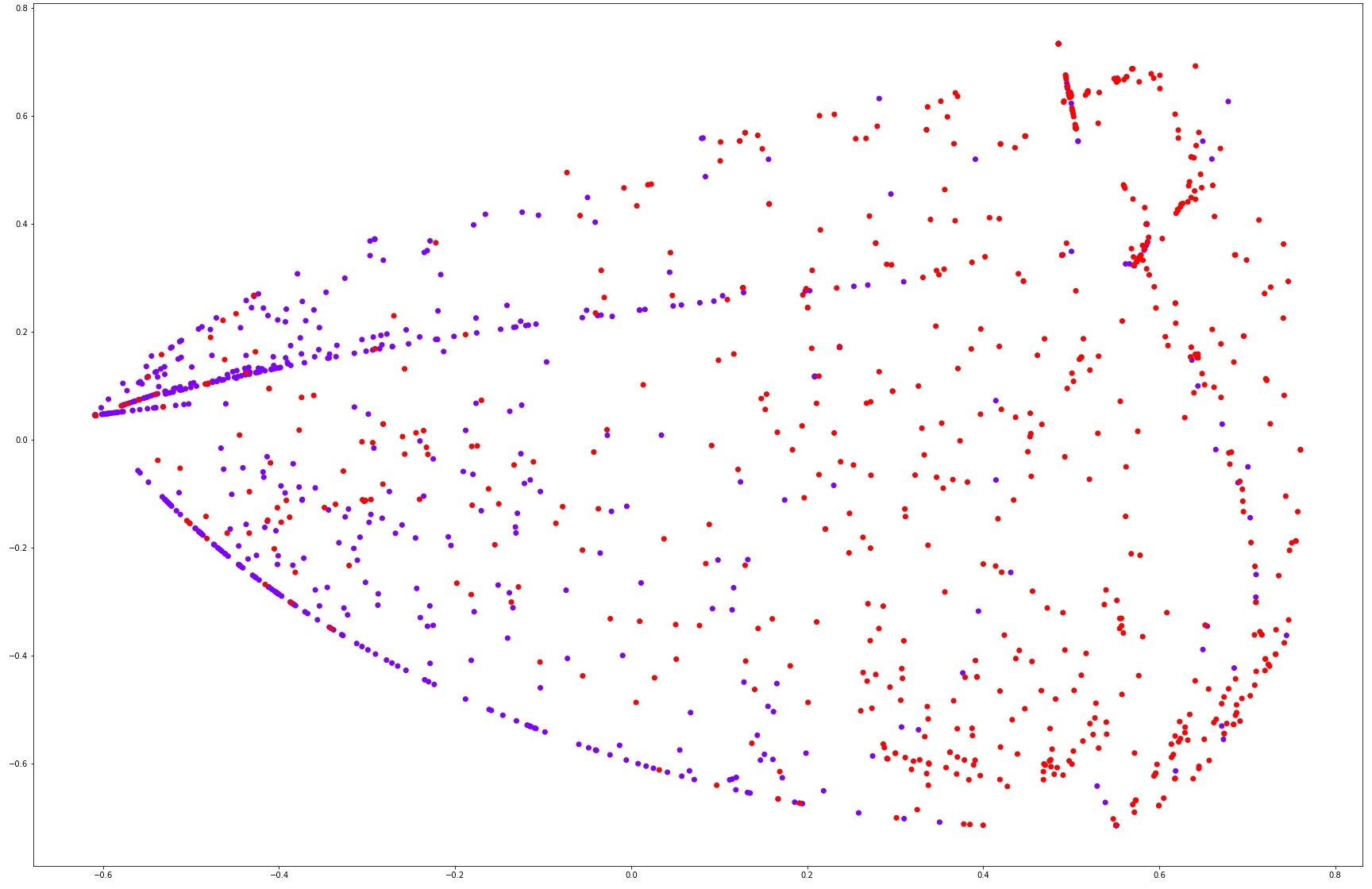}}&
		\subfloat[Prediction]{\includegraphics[width=0.45\textwidth]{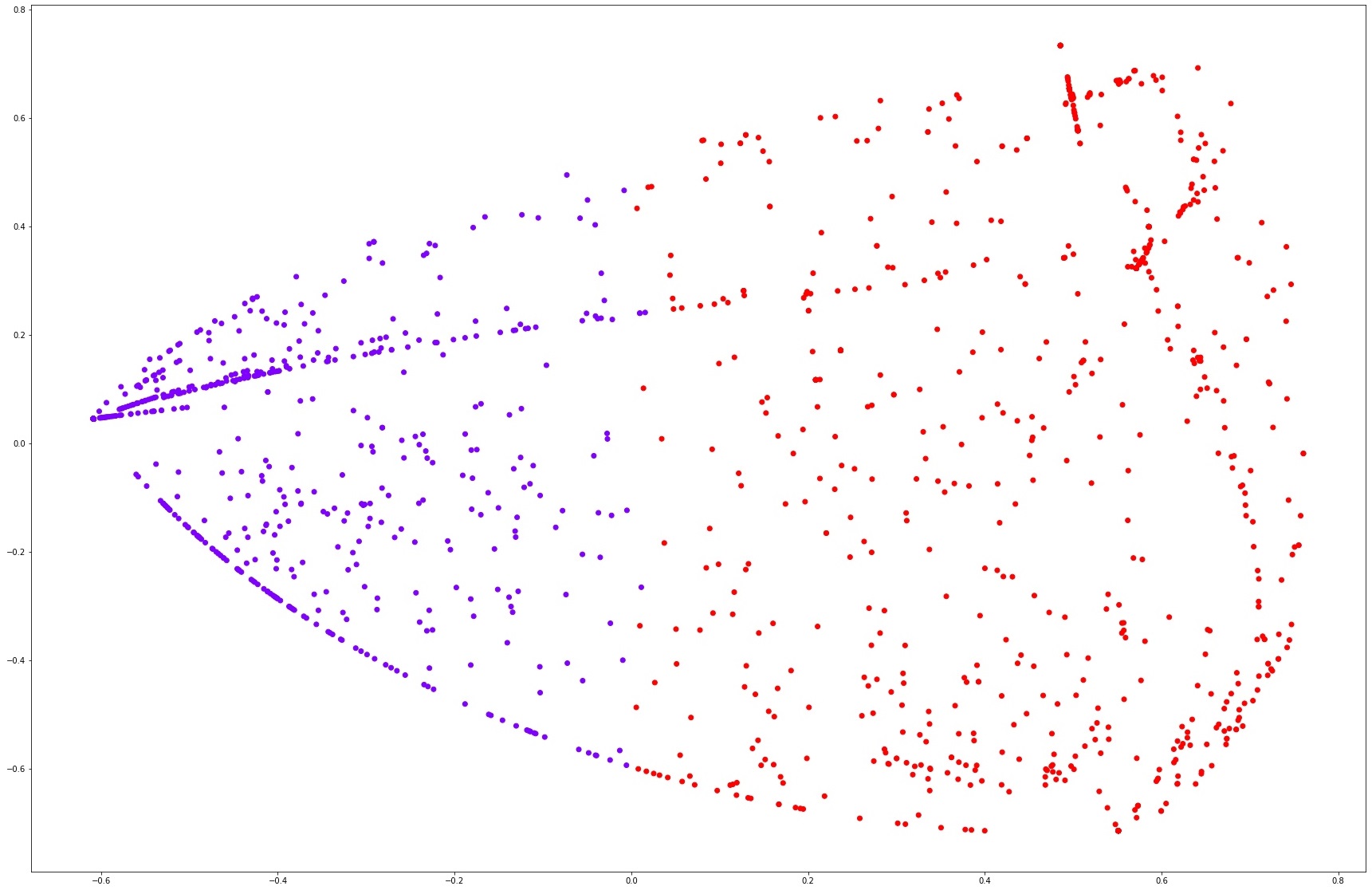}}
		\\
	\end{tabular}
	
	\caption{Comparison between the clustering of the original data (a) and (b), and the data generated using the proposed method (c) and (d).}
	\label{fig:pcacomparison2}
\end{figure*}

The proposed method's advantage over LSA is not only its stability or small improvement in clustering performance, but rather in the space that it generates. LSA basically creates a new space logically similar to the original, while the proposed method creates a more different space which is more suitable for clustering. An instance of the spaces generated by the original term-document matrix obtained from the BBC News dataset \cite{greene2006practical}, the LSA reduction and the proposed method's new feature space is provided in Fig. \ref{fig:pcacomparison1}. We should note that the datasets presented in this figure were projected to 2-dimensional space using PCA. Moreover, an instance of the SMS Spam Collection dataset \cite{almeida2011} being clustered using the original data vs the proposed method is presented in Fig.
\ref{fig:pcacomparison2}.

\subsection{Statistical Analysis of the results}
In this subsection, we present the results of the Wilcoxon Signed Ranks test \cite{wilcoxon1945individual} on the results from the previous subsections. The results are provided below in Table \ref{tab:wilcoxon}. The values under the 0.05 threshold are emboldened in the table. As it can be observed, the proposed method shows significant improvement over the clustering methods with great certainty in all metrics. When it comes to LSA combined with K-Means however, the proposed method holds the same standard only in terms of accuracy and ARI, while significant improvement in the other three metrics cannot be inferred with great certainty.
\begin{table}[!h]
	\centering
	\caption{Asymptotic p-values obtained from the Wilcoxon test comparing the proposed method to other methods.}
	\label{tab:wilcoxon}
	\begin{tabular}{p{0.1\textwidth}p{0.125\textwidth}p{0.125\textwidth}p{0.125\textwidth}}
		\hline\noalign{\smallskip}
		Method&Accuracy&NMI&ARI\\
		\noalign{\smallskip}\hline\noalign{\smallskip}
		
		GAKM &\textbf{0.001944}&\textbf{0.001944}&\textbf{0.001944}\\
		KM   &\textbf{0.001944}&\textbf{0.001944}&\textbf{0.001944}\\
		SKM  &\textbf{0.001944}&\textbf{0.004193}&\textbf{0.001944}\\
		SCPSO&\textbf{0.001944}&\textbf{0.004193}&\textbf{0.002526}\\
		LSAKM&\textbf{0.013471}&0.077556&\textbf{0.020658}\\
		
		\noalign{\smallskip}\hline
	\end{tabular}
\end{table}

\subsection{Robustness analysis}
In this subsection, we present the robustness analysis of the proposed method. Due to the NMF initialization using Singular Value Decomposition, the likeliness of NMF optimization reaching very similar results is very high. Following that, the proposed approach uses NMF to group features and therefore discretizes the components matrix instead of using direct output. As a result, the new feature space is likely to be very stable. Moreover, we present the standard deviation of the results of all of the compared methods in Tables \ref{tab:std_acc} - \ref{tab:std_ari}.

\begin{table}[!h]
	\centering
	\caption{Comparison of the standard deviation over accuracy}
	\label{tab:std_acc}
	\begin{tabular}{p{0.11\textwidth}p{0.11\textwidth}p{0.11\textwidth}p{0.11\textwidth}p{0.11\textwidth}p{0.11\textwidth}p{0.11\textwidth}}
		\hline\noalign{\smallskip}
		Dataset&GAKM&KM&LSAKM&SCPSO&SKM&NMF-FR\\
		\noalign{\smallskip}\hline\noalign{\smallskip}
		
		20COMP&0.034798&0.025838&0.000671&0.00143&0.032201&\textbf{0}\\ 
		20POL&0.052603&0.034765&0.05105&0.029353&0.015905&\textbf{0}\\ 
		20MISC&0.093977&0.022207&0.085553&0.001327&0.033934&\textbf{0}\\ 
		20REL&0.039602&0.011993&0.004374&0.008986&0.017193&\textbf{0}\\ 
		20SCI&0.071488&0.042292&0.059205&0.005267&0.059754&\textbf{0}\\ 
		AG&0.02649&0.040021&0.064437&0.048045&0.026684&\textbf{0}\\ 
		BBCSP&0.056641&0.069367&0.042925&0.042471&0.101892&\textbf{1.11E-16}\\ 
		BBC&0.07544&0.07324&0.000461&0.026571&0.072749&\textbf{1.11E-16}\\ 
		DMOZ&0.014411&0.028526&0.014379&0.051547&0.01883&\textbf{0}\\ 
		SMS&0.036645&0.091285&0.000325&0.007021&0.048861&\textbf{0}\\ 
		WEB&0.029667&0.030231&0.031363&0.042913&0.027141&\textbf{0}\\ 
		WEBACE&0.003493&0.015729&0.008388&0.013249&0.030847&\textbf{0}\\ 
		
		\noalign{\smallskip}\hline
	\end{tabular}
\end{table}

\begin{table}[!h]
	\centering
	\caption{Comparison of the standard deviation over NMI}
	\label{tab:std_nmi}
	\begin{tabular}{p{0.11\textwidth}p{0.11\textwidth}p{0.11\textwidth}p{0.11\textwidth}p{0.11\textwidth}p{0.11\textwidth}p{0.11\textwidth}}
		\hline\noalign{\smallskip}
		Dataset&GAKM&KM&LSAKM&SCPSO&SKM&NMF-FR\\
		\noalign{\smallskip}\hline\noalign{\smallskip}
		
		20COMP&0.023827&0.01672&0.00062&0.000509&0.03523&\textbf{0}\\ 
		20POL&0.058034&0.065036&0.075231&0.057869&0.029179&\textbf{0}\\ 
		20MISC&0.117595&0.030456&0.055941&0.000951&0.041503&\textbf{0}\\ 
		20REL&0.014619&0.011882&0.000861&0.004575&0.011412&\textbf{0}\\ 
		20SCI&0.055601&0.036521&0.05377&0.025587&0.067244&\textbf{0}\\ 
		AG&0.028049&0.020567&0.058411&0.03752&0.042798&\textbf{0}\\ 
		BBCSP&0.065146&0.089037&0.046097&0.048137&0.079677&\textbf{0}\\ 
		BBC&0.050704&0.061052&0.001015&0.046605&0.059978&\textbf{0}\\ 
		DMOZ&0.011902&0.025437&0.004994&0.032689&0.018252&\textbf{0}\\ 
		SMS&0.040704&0.130669&0.000512&0.02541&0.050606&\textbf{0}\\ 
		WEB&0.034775&0.03863&0.015435&0.028148&0.018852&\textbf{0}\\ 
		WEBACE&0.007109&0.011948&0.004956&0.009305&0.023649&\textbf{0}\\

		\noalign{\smallskip}\hline
	\end{tabular}
\end{table}

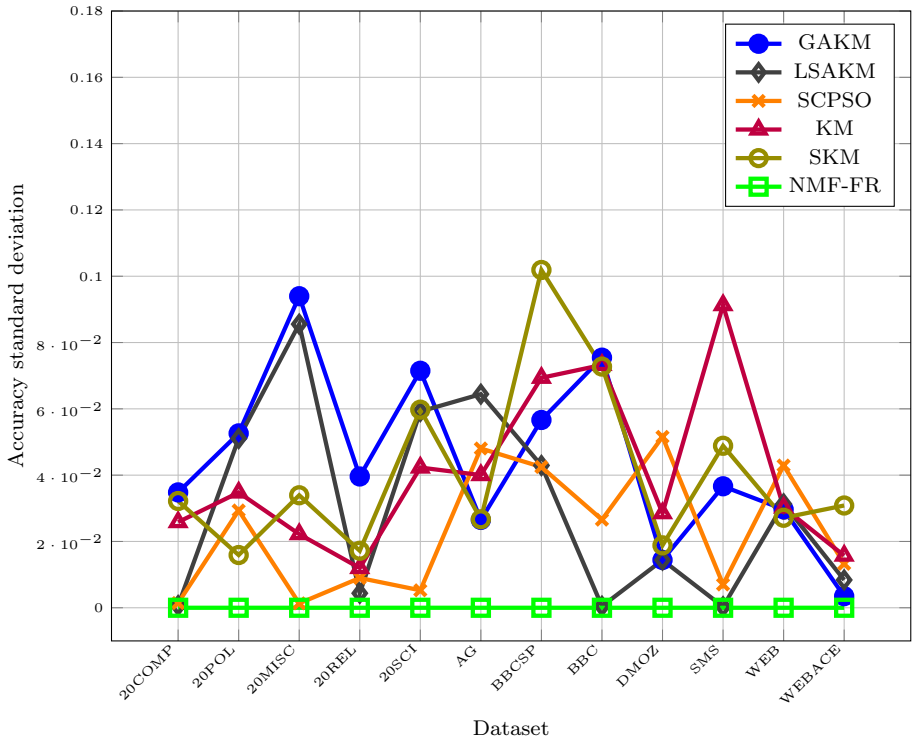
\begin{figure}[h]
	\centering
	\begin{tikzpicture}
	\begin{axis}[
	width=0.99\textwidth,
	height=0.5\textheight,
	grid=both,
	xlabel={Dataset},
	ylabel={Accuracy standard deviation},
	tick label style={font=\tiny},
	ymin=-0.01, ymax=0.18,
	xtick=data,
	x tick label style={rotate=45,anchor=east},
	symbolic x coords={20COMP,20POL,20MISC,20REL,20SCI,AG,BBCSP,BBC,DMOZ,SMS,WEB,WEBACE},
	point meta=explicit symbolic,
	x label style={at={(axis description cs:0.5,-0.05)},anchor=north},
	]
	
	\addplot[
	color=blue,
	mark=*,
	ultra thick,
	mark size=3pt,
	]
	coordinates {
		(20COMP,0.034798)(20POL,0.052603)(20MISC,0.093977)(20REL,0.039602)(20SCI,0.071488)(AG,0.02649)(BBCSP,0.056641)(BBC,0.07544)(DMOZ,0.014411)(SMS,0.036645)(WEB,0.029667)(WEBACE,0.003493)
	};
	\addlegendentry{GAKM}
	
	\addplot[
	color=darkgray,
	mark=diamond,
	ultra thick,
	mark size=3pt,
	]
	coordinates {
		(20COMP,0.000671)(20POL,0.05105)(20MISC,0.085553)(20REL,0.004374)(20SCI,0.059205)(AG,0.064437)(BBCSP,0.042925)(BBC,0.000461)(DMOZ,0.014379)(SMS,0.000325)(WEB,0.031363)(WEBACE,0.008388)
	};
	\addlegendentry{LSAKM}
	
	\addplot[
	color=orange,
	mark=x,
	ultra thick,
	mark size=3pt,
	]
	coordinates {
		(20COMP,0.00143)(20POL,0.029353)(20MISC,0.001327)(20REL,0.008986)(20SCI,0.005267)(AG,0.048045)(BBCSP,0.042471)(BBC,0.026571)(DMOZ,0.051547)(SMS,0.007021)(WEB,0.042913)(WEBACE,0.013249)
	};
	\addlegendentry{SCPSO}
	
	\addplot[
	color=purple,
	mark=triangle,
	ultra thick,
	mark size=3pt,
	]
	coordinates {
		(20COMP,0.025838)(20POL,0.034765)(20MISC,0.022207)(20REL,0.011993)(20SCI,0.042292)(AG,0.040021)(BBCSP,0.069367)(BBC,0.07324)(DMOZ,0.028526)(SMS,0.091285)(WEB,0.030231)(WEBACE,0.015729)
	};
	\addlegendentry{KM}
	
	\addplot[
	color=olive,
	mark=o,
	ultra thick,
	mark size=3pt,
	]
	coordinates {
		(20COMP,0.032201)(20POL,0.015905)(20MISC,0.033934)(20REL,0.017193)(20SCI,0.059754)(AG,0.026684)(BBCSP,0.101892)(BBC,0.072749)(DMOZ,0.01883)(SMS,0.048861)(WEB,0.027141)(WEBACE,0.030847)
	};
	\addlegendentry{SKM}
	
	\addplot[
	color=green,
	mark=square,
	ultra thick,
	mark size=3pt,
	]
	coordinates {
		(20COMP,0)(20POL,0)(20MISC,0)(20REL,0)(20SCI,0)(AG,0)(BBCSP,1.11E-16)(BBC,1.11E-16)(DMOZ,0)(SMS,0)(WEB,0)(WEBACE,0)
	};
	\addlegendentry{NMF-FR}

	\end{axis}
	\end{tikzpicture}
	\caption{Accuracy standard deviation of the proposed method and other methods.}
	\label{fig:robustness}
\end{figure}

\begin{table}[!h]
	\centering
	\caption{Comparison of the standard deviation over ARI}
	\label{tab:std_ari}
	\begin{tabular}{p{0.11\textwidth}p{0.11\textwidth}p{0.11\textwidth}p{0.11\textwidth}p{0.11\textwidth}p{0.11\textwidth}p{0.11\textwidth}}
		\hline\noalign{\smallskip}
		Dataset&GAKM&KM&LSAKM&SCPSO&SKM&NMF-FR\\
		\noalign{\smallskip}\hline\noalign{\smallskip}
		
		20COMP&0.026859&0.018394&0.000854&5.83E-05&0.02786&\textbf{2.78E-17}\\ 
		20POL&0.065709&0.037183&0.057482&0.049329&0.04802&\textbf{0}\\ 
		20MISC&0.058746&0.031436&0.062107&0.000224&0.035744&\textbf{0}\\ 
		20REL&0.025577&0.00656&0.001664&0.017934&0.007629&\textbf{0}\\ 
		20SCI&0.054933&0.033139&0.063391&0.031548&0.061382&\textbf{0}\\ 
		AG&0.019696&0.039829&0.068032&0.075245&0.024156&\textbf{0}\\ 
		BBCSP&0.055734&0.11875&0.048808&0.084792&0.163983&\textbf{0}\\ 
		BBC&0.085176&0.10793&0.00105&0.04449&0.101081&\textbf{0}\\ 
		DMOZ&0.007424&0.017467&0.009887&0.048212&0.021862&\textbf{0}\\ 
		SMS&0.052473&0.147469&0.000769&0.003445&0.086976&\textbf{0}\\ 
		WEB&0.041166&0.032724&0.025355&0.031392&0.026607&\textbf{0}\\ 
		WEBACE&0.005566&0.025055&0.008275&0.016236&0.0226&\textbf{0}\\

		\noalign{\smallskip}\hline
	\end{tabular}
\end{table}
Once again, it can be observed, the proposed method has a standard deviation of zero (or in some cases an infinitesimally small deviation) over multiple runs, when compared to the rest. Moreover, we present the accuracy standard deviation in Fig. \ref{fig:robustness}.

\section{Conclusion}
\label{sec:conclusion}
As we mentioned, text mining plays a very crucial part in many computerized systems nowadays, such as web searches, recommendation systems and the like. Advances in this area of research, specifically in text clustering includes but is not limited to matrix analysis and specific clustering methods.

In this paper, we propose a new dimension reduction method based on Nonnegative Matrix Factorization, which can be used to group the terms obtained from the term-document matrix. Afterwards, the method agglomerates each group's features into one new feature vector, by taking their norm values. Therefore, a number of n-dimensional feature vectors (n being the number of documents) are combined into a single n-dimensional feature vector. The new feature vectors, which together create a new data matrix can then be further reduced using LSA. This newly generated space is more suitable for clustering than the original.
Afterwards, spherical K-Means is used to cluster the new feature space, which is initialized using a new approach which chooses the densest areas of the space as the initial centroids through creating a nearest-neighbors graph.
In the proposed method, we initialize NMF using Singular Value Decomposition, which will increase the robustness of the method. Because of the coordinate-based optimization of NMF, even slight differences in SVD will lead to similar results from NMF. The output components matrix from NMF is then used to separate terms into groups. Even slight changes in this matrix will not lead to different feature outputs produced by the proposed feature agglomeration since the components matrix is discretized in order to group the terms. Therefore, the robustness of the results produced from the agglomeration is very high and therefore the method is positively stable. This stability is followed by a non-stochastic initialization of K-Means, which makes the proposed method very deterministic.
The deterministic initialization operates by creating a nearest-neighbors graph of the newly generated space, and selecting the most-connected records as the initial centroids. This supports faster convergence in K-Means, and has proved to be approximately as good as K-Means++, while being deterministic in nature, as opposed to K-Means++.
We conducted experiments on 12 text classification datasets and inspected external clustering evaluation measures, namely Purity, Normalized Mutual Information, and Adjusted Rand Index. We also compared the proposed method to two of the recently proposed methods, as well as two classical clustering approaches. The results showed significant improvement to most, while showing not necessarily significant improvement to spherical K-Means in terms of clustering scores. We also compared the proposed method to LSA-based Spherical K-Means and also showed improvement in clustering scores. Nevertheless, the most important advantage of the proposed method is its stability in results which is also presented. The proposed method reaches a standard deviation of zero in most cases, while reaching near-zero values in others.
The proposed method can also be further improved in the future by possibly exploring other clustering measures, using measures other than the L2 norm to agglomerate features as well as other matrix factorization methods instead of NMF. Another area which can be explored in the future is using other NMF initializers instead of SVD.

\section*{Conflict of interest}
The authors declare that they have no conflict of interest.

\bibliographystyle{unsrtspbasic}      
\bibliography{references}   

\end{document}